\newif\ifconfver
\confvertrue        

\newif\ifonecoltab
\onecoltabtrue        

\newif\ifplainver  
\plainvertrue

\ifplainver
    \confverfalse                         
\fi

\ifconfver
     \documentclass[10pt,journal]{IEEEtran}
\else
    \ifplainver
        \documentclass[11pt]{article}
        \usepackage{fullpage}
    \else
        \documentclass[12pt,draftcls,onecolumn]{IEEEtran}
    \fi
\fi

\usepackage{calc,amsfonts,amssymb,amsmath,bm,url,color,theorem,graphicx,cite,epstopdf,nicefrac}
\usepackage{psfrag,subfigure,float}
\usepackage[algoruled,linesnumbered]{algorithm2e}
\usepackage{multirow}
\usepackage[normalem]{ulem}

\definecolor{orange}{RGB}{255,107,0}

\newtheorem{Lemma}{Lemma}
\newtheorem{Prop}{Proposition}
\newtheorem{Theorem}{Theorem}

\theorembodyfont{\rmfamily}

\newtheorem{Remark}{Remark}

\newcommand{\G}{\boldsymbol{G}}
\newcommand{\Q}{\boldsymbol{Q}}
\newcommand{\X}{\boldsymbol{X}}

\newcommand{\U}{\boldsymbol{U}}

\newcommand{\M}{\boldsymbol{M}}

\begin{document}

\newcommand{\papertitle}{
Scalable and Flexible Multiview MAX-VAR Canonical Correlation Analysis
}

\newcommand{\paperabstract}{
Generalized canonical correlation analysis (GCCA) aims at finding latent low-dimensional common structure from multiple views (feature vectors in different domains) of the same entities.
Unlike principal component analysis (PCA) that handles a single view, (G)CCA is able to integrate information from different feature spaces. Here we focus on MAX-VAR GCCA, a popular formulation which has recently
gained renewed interest in multilingual processing and speech modeling.
The classic MAX-VAR GCCA problem can be solved optimally via eigen-decomposition of a matrix that compounds the (whitened) correlation matrices of the views; but this solution has serious scalability issues, and is not directly amenable to incorporating pertinent structural constraints such as non-negativity and sparsity on the canonical components.
We posit regularized MAX-VAR GCCA as a non-convex optimization problem and propose an alternating optimization (AO)-based algorithm to handle it. Our algorithm alternates between {\em inexact} solutions of a regularized least squares subproblem and a manifold-constrained non-convex subproblem, thereby achieving substantial memory and computational savings. An important benefit of our design is that it can easily handle structure-promoting regularization.
We show that the algorithm globally converges to a critical point at a sublinear rate, and approaches a global optimal solution at a linear rate when no regularization is considered. Judiciously designed simulations and large-scale word embedding tasks are employed to showcase the effectiveness of the proposed algorithm.
}


\ifplainver

    \date{\today}

    \title{\papertitle}

    \author{
    Xiao Fu$^\ast$, Kejun Huang$^\ast$, Mingyi Hong	$^\ddag$, Nicholas D. Sidiropoulos$^\ast$, and Anthony Man-Cho So$^\dag$
    \\ ~ \\
		$^\ast$Department of Electrical and Computer Engineering, University of Minnesota,\\
		Minneapolis, 55455, MN, United States\\
		Email: (xfu,huang663,nikos)@umn.edu
	  \\~\\
	$^\ddag$Department of Industrial and Manufacturing Systems Engineering, Iowa State University\\
	Ames, Iowa 50011, (515) 294-4111,\\
	Email: mingyi@iastate.edu
	 \\ ~ \\
    $^\dag$Department of Systems Engineering and Engineering Management\\ The Chinese University of Hong Kong, \\
	Shatin, N.T., Hong Kong \\
	Email: manchoso@se.cuhk.edu.hk
	\\
    }

    \maketitle

    \begin{abstract}
    \paperabstract
    \end{abstract}

\else
    \title{\papertitle}

    \ifconfver \else {\linespread{1.1} \rm \fi

\author{Xiao Fu, \IEEEmembership{Member, IEEE}, Kejun Huang, \IEEEmembership{Member, IEEE}, Mingyi Hong, \IEEEmembership{Member, IEEE}\\Nicholas D. Sidiropoulos, \IEEEmembership{Fellow, IEEE}, and Anthony Man-Cho So, \IEEEmembership{Member, IEEE}
\thanks{X. Fu, K. Huang and N.D. Sidiropoulos are supported in part by National Science Foundation under Project NSF ECCS-1608961 and Project NSF IIS-1447788.
M. Hong is supported in part by National Science Foundation under Project NSF CCF-1526078 and by the Air Force Office of Scientific Research (AFOSR) under Grant 15RT0767.
	
X. Fu, K. Huang and N.D. Sidiropoulos are with the Department of Electrical and Computer Engineering, University of Minnesota, Minneapolis, MN 55455, e-mail (xfu,huang663,nikos)@umn.edu.

M. Hong is with Department of Industrial and Manufacturing Systems Engineering, Iowa State University,	Ames, Iowa 50011, (515) 294-4111, Email: mingyi@iastate.edu.

Anthony M.-C. So is with the Department of Systems Engineering and Engineering Management, The Chinese University of Hong Kong, Shatin, N.T., Hong Kong, Email: manchoso@se.cuhk.edu.hk
}
}

    \maketitle

    \ifconfver \else
        \begin{center} \vspace*{-2\baselineskip}
        \end{center}
    \fi

    \begin{abstract}
    \paperabstract
    \end{abstract}

    \begin{keywords}\vspace{-0.0cm}
        Canonical correlation analysis (CCA), multiview CCA, MAX-VAR, word embedding, optimization, scalability, feature selection
    \end{keywords}

    \ifconfver \else \IEEEpeerreviewmaketitle} \fi

 \fi

\ifconfver \else
    \ifplainver \else
        \newpage
\fi \fi

\section{Introduction}
\begin{figure}[t]
\centering
\includegraphics[width=.55\linewidth]{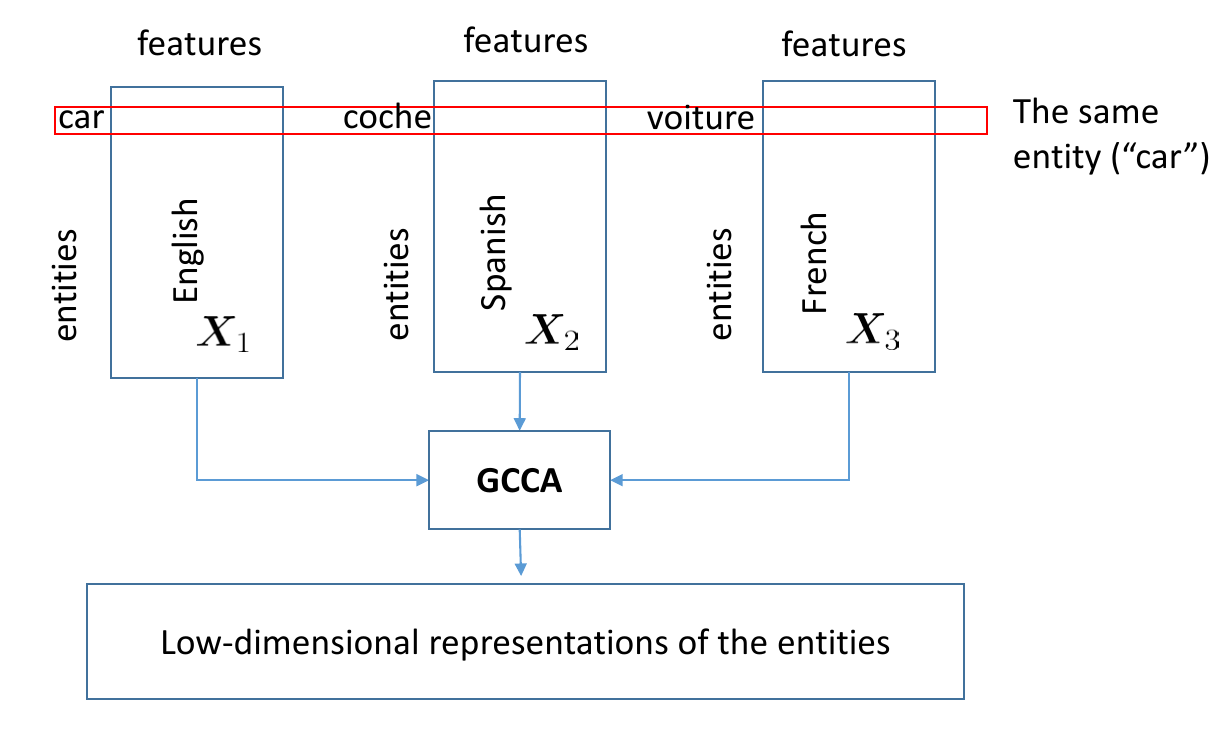}
\caption{Word embedding seeks low-dimensional representations of the entities (words) that are well-aligned with human judgment. Different language data (i.e., ${\bm X}_1$-$\X_3$) can be considered as different views / feature spaces of the same entities.}
\label{fig:motivation}
\end{figure}

Canonical correlation analysis (CCA) \cite{hardoon2004canonical} produces low-dimensional representations via finding common structure of two or more views corresponding to the same entities.
A view contains high-dimensional representations of the entities in a certain feature space -- e.g., the text and audio representations of a given word can be considered as different views of this word.
CCA is able to deal with views that have different dimensions, and this flexibility is very useful in data fusion, where one is interested in integrating information acquired from different domains.
Multiview analysis finds numerous applications in signal processing and machine learning, such as
blind source separation \cite{li2009joint,bertrand2015distributed}, direction-of-arrival estimation \cite{wu1994music}, wireless channel equalization \cite{dogandzic2002finite}, regression \cite{kakade2007multi}, clustering \cite{chaudhuri2009multi}, speech modeling and recognition \cite{arora2014multi,wang2015acoustic}, and word embedding \cite{rastogimultiview}, to name a few.
Classical CCA was derived for the two-view case, but \textit{generalized canonical correlation analysis} (GCCA) that aims at handling more than two views has a long history as well \cite{carroll1968generalization,horst1961generalized}.
A typical application of GCCA, namely, multilingual word embedding, is shown in  Fig.~\ref{fig:motivation}.
Applying GCCA to integrate multiple languages was shown to yield better embedding results relative to single-view analyses such as principle component analysis (PCA) \cite{rastogimultiview}.


Computationally, GCCA poses interesting and challenging optimization problems.
Unlike the two-view case that admits an algebraically simple solution (via eigen-decomposition),
GCCA is in general not easily solvable.
Many prior works considered the GCCA problem with different cost functions \cite{carroll1968generalization,kettenring1971canonical,horst1961generalized} -- see a nice summary in \cite[Chapter 10]{asendorf2015informative}. However, the proposed algorithms often can only extract a single canonical component and then find others through a deflation process, which is known to suffer from error propagation.
CCA and GCCA can also pose serious scalability challenges,
since they involve auto- and/or cross-correlations of different views and a whitening stage \cite{ma2015finding}.
These procedures can easily lead to memory explosion and require a large number of flops for computation.
They also destroy the sparsity of the data, which is usually what one relies upon to deal with large-scale problems.
In recent years, effort has been spent on solving these scalability issues, but the focus is mostly on the two-view case \cite{ma2015finding,sun2011canonical,lu2014large}.

Among all different formulations of GCCA, there is a particular one that admits a conceptually simple solution, the so-called MAX-VAR GCCA \cite{carroll1968generalization,van2006generalized,kettenring1971canonical}.
MAX-VAR GCCA was first proposed in \cite{horst1961generalized}, and its solution amounts to finding the `directions' aligned to those exhibiting maximum variance for a matrix aggregated from the (whitened) auto-correlations of the views.
It can also be viewed as a problem of enforcing {\em identical} latent representations of different views as opposed to highly correlated ones, which is the more general goal of (G)CCA.
The merit of MAX-VAR GCCA is that it can be solved via eigen-decomposition and finds all the canonical components simultaneously. 
In practice, MAX-VAR GCCA also demonstrates promising performance in various applications such as word embedding \cite{rastogimultiview} and speech recognition \cite{arora2014multi}.
On the other hand, MAX-VAR GCCA has the same scalability problem as the other GCCA formulations:
It involves correlation matrices of different views and their inverses, which is prohibitive to even instantiate when the data dimension is large.
The work in \cite{rastogimultiview} provided a pragmatic way to circumvent this difficulty: PCA was first applied to each view to reduce the rank of the views,
and then MAX-VAR GCCA was applied to the rank-truncated views. Such a procedure significantly reduces
the number of parameters for characterizing the views and is feasible in terms of memory. However, truncating the rank of the views is prone to information loss, and thus leads to performance degradation.

Besides the basic (G)CCA formulations, \emph{structured} (G)CCA \cite{hardoon2011sparse} that
seeks canonical components with pre-specified structure is often considered in applications.
Sparse/group-sparse CCA has attracted particular attention, since it has the ability of discarding outlying or irrelevant features when performing CCA \cite{witten2009penalized,chen2012structured,witten2009extensions}.
In multi-lingual word embedding \cite{faruqui2014improving,sun2011canonical,rastogimultiview}, for example, it is known that outlying features (``stop words''), may exist.
Gene analysis is another example \cite{witten2009penalized,chen2012structured,witten2009extensions}.
Ideally, CCA seeks a few highly correlated latent components, and so it should naturally be able to identify and down-weight irrelevant features automatically. In practice, however, this ability is often impaired when correlations cannot be reliably estimated, when one only has access to relatively few and/or very noisy samples, or when there is model mismatch due to bad preprocessing. In those cases, performing feature selection jointly with (G)CCA is well-motivated.
Some other structure-promoting regularizations may also be of interest: Non-negativity together with sparsity have proven  helpful in analyzing audio and video data, since non-negative CCA produces weighted sums of video frames that are interpretable \cite{sigg2007nonnegative};
non-negative CCA has also proven useful in time series analysis \cite{fischer2007time}.
Effective algorithms that tackle large-scale structured GCCA problems are currently missing, to the best of our knowledge. 

\smallskip

\noindent
{\bf Contributions}
In this work,
our goal is to provide a scalable and flexible algorithmic framework for handling
the MAX-VAR GCCA problem and its variants with structure-promoting regularizers.
Instead of truncating the rank of the views as in \cite{rastogimultiview}, we keep the data \emph{intact} and deal with the problem using a two-block alternating optimization (AO) framework. The proposed algorithm alternates between a regularized least squares subproblem and an orthogonality-constrained subproblem.
The merit of this framework is that correlation matrices of the views never need to be explicitly instantiated, and the inversion procedure is avoided.
Consequently, the algorithm consumes significantly less memory compared to that required by the original solution using eigen-decomposition.
The proposed algorithm allows inexact solution to the subproblems,
and thus per-iteration computational complexity is also light.
In addition, it can easily handle
different structure-promoting regularizers (e.g. sparsity, group sparsity and non-negativity) without increasing memory and computational costs, including the feature-selective regularizers that we
are mainly interested in.

The AO algorithm alternates between convex and non-convex manifold-constrained subproblems, using possibly inexact updates for the subproblems. Under such circumstances, general
convergence analysis tools cannot be directly applied, and thus the associated convergence properties are not obvious. This necessitates custom convergence analysis. We first show that the proposed algorithm \emph{globally} converges to a Karush-Kuhn-Tucker (KKT) point of the formulated problem, even when a variety of regularizers are employed.
We also show that the optimality gap shrinks to at most ${\cal O}(1/r)$ after $r$ iterations -- i.e.,
at least a sublinear convergence rate can be guaranteed.
In addition, we show that when the classic MAX-VAR problem without regularization (or with a minimal energy regularization) is considered, the proposed algorithm approaches a \textit{global optimal solution} and enjoys a \textit{linear} convergence rate.

The proposed algorithm is applied to judiciously designed simulated data, as well as a real large-scale word embedding problem, and promising results are observed.

A conference version of this work appears at ICASSP 2017, New Orleans, USA, Mar. 2017 \cite{fu2016scalable}. This journal version includes detailed convergence analysis and proofs, comprehensive simulations, and a set of experiments using real large-scale multilingual data.

\smallskip

\noindent
{\bf Notation} We use $\X$ and ${\bm x}$ to denote a matrix and a vector, respectively.
$\X(m,:)$ and $\X(:,n)$ denote the $m$th row and the $n$th column of $\X$, respectively;
in particular, $\X(:,n_1:n_2)$ ($\X(n_1:n_2,:)$) denotes a submatrix of $\X$ consisting of the $n_1$-$n_2$th columns (rows) of $\X$ (\texttt{MATLAB} notation).
$\|\X\|_F$ and $\|\X\|_p$ for $p\geq 1$ denote the Frobenius norm and the matrix-induced $p$-norm, respectively. $\|\X\|_{p,1} = \sum_{i=1}^m \|\X(i,:)\|_p$ for $p\geq 1$ denotes the $\ell_p/\ell_1$-mixed norm of $\X\in\mathbb{R}^{m\times n}$.
The superscripts ``$T$'', ``$\dag$'', and ``${-1}$'' denote the matrix operators of transpose, pseudo-inverse and inverse, respectively.
The operator $\left<{\bm X},{\bm Y}\right>$ denotes the inner product of ${\bm X}$ and ${\bm Y}$.
${\bm 1}_{+}({\bm X})$ denotes the element-wise indicator function of the nonnegative orthant -- i.e.,
${\bm 1}_{+}({\bm X})=+\infty$ if any element of ${\bm X}$ is negative and ${\bm 1}_{+}({\bm X})=0$ otherwise.

\vspace{-.15cm}
\section{Background}

Consider a scenario where $L$ entities have different representations in $I$ views.
Let ${\bm X}_i\in\mathbb{R}^{L\times M_i}$ denote
the $i$th view with its $\ell$th row ${\bm X}_i(\ell,:)$ being a feature vector that defines the $\ell$th data point (entity) in the $i$th view (cf. Fig.~\ref{fig:motivation}), where $M_i$ is the dimension of the $i$th feature space.
The classic two-view CCA aims at finding common structure of the views via linear transformation.
Specifically, the corresponding problem can be expressed in the following form \cite{hardoon2004canonical}:
\begin{subequations}\label{eq:CCA}
\begin{align}
    \min_{{\bm Q}_1,{\bm Q}_2} &~\left\|{\bm X}_1{\bm Q}_1-{\bm X}_2{\bm Q}_2\right\|_F^2\\
                                  {\rm s.t.} &~{\bm Q}_i^T\left({\bm X}_i^T{\bm X}_i\right){\bm Q}_i={\bm I},\quad i=1,2, \label{eq:normalization}
\end{align}
\end{subequations}
where the columns of ${\bm Q}_i\in\mathbb{R}^{M_i\times K}$ correspond to the $K$ canonical components of view ${\bm X}_i$, and $K$ is usually small (i.e., $K\ll \min\{M_i,L\}$).
Note that we are essentially maximizing the trace of the estimated cross-correlations between the reduced-dimension views, i.e., ${\rm Tr}( \bm{Q}_2^T\bm{X}_2^T\bm{X}_1\bm{Q}_1)$ subject to the normalization in \eqref{eq:normalization} -- which motivates the terminology ``correlation analysis''.
Problem~\eqref{eq:CCA} can be solved via a generalized eigen-decomposition, but this simple solution only applies to the two-view case.
To analyze the case with more than two views,
one natural thought is to extend the formulation in \eqref{eq:CCA} to a pairwise matching cirterion, i.e., $\sum_{i=1}^{I-1}\sum_{j=i+1}^I\left\|{\bm X}_i{\bm Q}_i-{\bm X}_j{\bm Q}_j\right\|_F^2$ with orthogonality constraints on ${\bm X}_i{\bm Q}_i$ for all $i$, where $I$ is the number of views.
Such an extension leads to the so-called sum-of-correlations (SUMCOR) generalized CCA \cite{carroll1968generalization}, which has been shown to be NP-hard \cite{rupnik2013comparison}. Notice that designing efficient and scalable algorithms for SUMCOR is an interesting topic and it started attracting attention recently \cite{rupnik2013comparison,zhang2011towards,fu2016efficient}.
Another formulation of GCCA is more tractable: Instead of forcing pairwise similarity of the reduced-dimension views,
one can seek a common latent representation of different views, i.e., \cite{carroll1968generalization,rastogimultiview,arora2014multi,van2006generalized,kettenring1971canonical}
\begin{equation}\label{eq:CGCCA}
\begin{aligned}
\min_{\{{\bm Q}_i\}_{i=1}^I,{\bm G}} &~\sum_{i=1}^{I}(\nicefrac{1}{2})\left\|{\bm X}_i{\bm Q}_i-{\bm G}\right\|_F^2,\\
                {\rm s.t.}&~{\bm G}^T{\bm G}={\bm I},
\end{aligned}
\end{equation}
where ${\bm G}\in\mathbb{R}^{L\times K}$ is a common latent representation of the different views.
Problems~\eqref{eq:CGCCA} also finds highly correlated reduced-dimension views as SUMCOR does.
The upshot of Problem~\eqref{eq:CGCCA} is that it ``transfers'' the multiple difficult constraints ${\bm Q}_i^T\X_i^T\X_i\Q_i={\bm I}$ to a single constraint $\G^T\G={\bm I}$, and thus admits a {\em conceptually} simple algebraic solution, which, as we will show, has the potential to be scaled up to deal with very large problems.
In this work, we will focus on Problem~\eqref{eq:CGCCA} and its variants.

Problem \eqref{eq:CGCCA} is referred to as the MAX-VAR formulation of GCCA since the optimal solution amounts to taking principal eigenvectors of a matrix aggregated from the correlation matrices of the views.
To explain, let us first assume that ${\bm X}_i$ has full column rank
and solve \eqref{eq:CGCCA} with respect to (w.r.t.) $\Q_i$, i.e., ${\bm Q}_i={\bm X}_i^\dag{\bm G}$, where ${\bm X}_i^\dag=({\bm X}_i^T{\bm X}_i)^{-1}{\bm X}_i^T$. By substituting it back to \eqref{eq:CGCCA}, we see that an optimal solution ${\bm G}_{\rm opt}$ can be obtained via solving the following:
\begin{equation}
\begin{aligned}
{\bm G}_{\rm opt} = \arg\max_{{\bm G}^T{\bm G} = {\bm I}} ~ {\rm Tr}\left({\bm G}^T\left(\sum_{i=1}^I {\bm X}_i{\bm X}_i^\dag\right){\bm G}\right). \label{eq:new_obj}
\end{aligned}
\end{equation}
Let ${\bm M} = \sum_{i=1}^I {\bm X}_i{\bm X}_i^\dag$.
Then, an optimal solution is ${\bm G}_{\rm opt}={\bm U}_M(:,1:K)$,
i.e., the first $K$ principal eigenvectors of ${\bm M}$ \cite{GHGolub1996}.
Although Problem~\eqref{eq:CGCCA} admits a seemingly easy solution, implementing it in practice has two major challenges:

\smallskip

\noindent
1) {\bf Scalability Issues}:
Implementing the eigen-decomposition based solution for large-scale data is prohibitive.
As mentioned, instantiating ${\bm M} = \sum_{i=1}^I{\bm X}_i({\bm X}_i^T{\bm X}_i)^{-1}{\bm X}_i^T$ is not doable when $L$ and $M_i$'s are large. The matrix ${\bm M}$ is an $L\times L$ matrix. In applications like word embedding, $L$ and $M_i$ are the vocabulary size of a language and the number of features defining the terms, respectively, which can both easily exceed $100,000$. This means that the memory for simply instantiating ${\bm M}$ or $({\bm X}_i^T{\bm X}_i)^{-1}$ can reach 75GB.	
In addition, even if the views ${\bm X}_i$ are sparse, computing $({\bm X}_i^T{\bm X}_i)^{-1}$ will create large dense matrices and make it difficult to exploit sparsity in the subsequent processing.
To circumvent these difficulties, Rastogi \emph{et al.} \cite{rastogimultiview} proposed to first apply the singular value decomposition (SVD) to the views, i.e., ${\rm svd}({\bm X}_i)={\bm U}_i{\bm \Sigma}_i{\bm V}_i^T$, and then let
$\hat{\bm X}_i = {\bm U}_i(:,1:P){\bm \Sigma}_i(1:P,1:P)({\bm V}_i(:,1:P))^T \approx {\bm X}_i,$
where $P$ is much smaller than $M_i$ and $L$.
This procedure enables one to represent the views with significantly fewer parameters, i.e., $(L+M_i+1)P$ compared to $LM_i$,
and allows the original eigen-decomposition based solution to MAX-VAR GCCA to be applied; see more details in \cite{rastogimultiview}. The drawback, however, is also evident: The procedure truncates the rank of the views significantly (since in practice the views almost always have full column-rank, i.e., ${\rm rank}({\bm X}_i)=M_i$), and rank-truncation is prone to information losses. Therefore, it is much more appealing to deal with the intact views.

\noindent
2) {\bf Structure-Promoting}:
Another aspect that is under-addressed by existing approaches is how to incorporate regularizations on $\Q_i$ to multiview large-scale CCA. Note that finding structured $\Q_i$ is well-motivated in practice. Taking multilingual word embedding as an example, $\X_i(:,n)$ represents
the $n$th feature in language $i$, which is usually defined by the co-occurrence frequency of the words and feature $n$ (also a word in language $i$).
However, many features of $\X_i$ may not be informative (e.g., ``the'' and ``a'' in English) or not correlated to data in $\X_j$.
These \textit{irrelevant} or \emph{outlying features} could result in unsatisfactory performance of GCCA if  not taken into account.
Under such scenarios, a more appealing formulation may include a row-sparsity promoting regularization on ${\bm Q}_i$ so that some columns corresponding to the irrelevant features in $\X_i$ can be discounted/downweighted when seeking $\Q_i$.
Sparse (G)CCA is desired in a variety of applications such as gene analytics and fMRI prediction \cite{rustandi2009integrating,mitchell2008predicting,witten2009penalized,chen2012structured,witten2009extensions}.
Other structure such as nonnegativity of ${\bm Q}_i$ was also shown useful in data analytics for maintaining interpretability and enhancing performance; see \cite{sigg2007nonnegative,fischer2007time}.

%

\section{Proposed Algorithm}
In this work, we consider a scalable and flexible algorithmic framework for handling MAX-VAR GCCA and its variants with structure-promoting regularizers on $\Q_i$.
We aim at offering
simple solutions that are memory-efficient, admit light per-iteration complexity, and feature good convergence properties under certain mild conditions.
Specifically, we consider the following formulation:
\begin{equation}\label{eq:CGCCA_reg}
\begin{aligned}
\min_{\{{\bm Q}_i\},\G} &~\sum_{i=1}^{I}(\nicefrac{1}{2})\left\|{\bm X}_i{\bm Q}_i-{\bm G}\right\|_F^2+\sum_{i=1}^I  h_i\left(\Q_i\right),\\
{\rm s.t.}&~\G^T\G={\bm I},
\end{aligned}
\end{equation}
where $h_i(\cdot)$ is a regularizer that imposes a certain structure on $\Q_i$.
Popular regularizers include 
\begin{subequations}
	\begin{align}
	h_i(\Q_i)&=\nicefrac{\mu_i}{2}\cdot\|\Q_i\|_{F}^2,\\
	h_i(\Q_i)&=\mu_i\cdot\|\Q_i\|_{2,1},\\
	h_i(\Q_i)&=\mu_i\cdot\|\Q_i\|_{1,1},\\
	h_i(\Q_i)&=\nicefrac{\mu_i}{2}\cdot\|\Q_i\|_{F}^2 + \beta_i \cdot\|\Q_i\|_{2,1}, \label{eq:elastic1}\\
	h_i(\Q_i)&=\nicefrac{\mu_i}{2}\cdot\|\Q_i\|_{F}^2 + \beta_i \cdot\|\Q_i\|_{1,1}, \label{eq:elastic2}\\
	h_i(\Q_i)&={\bm 1}_{+}({\bm Q}_i),
	\end{align}
\end{subequations}
where $\mu_i,\beta_i\geq 0$ are regularization parameters for balancing the least squares fitting term and the regularization terms.
The first regularizer is commonly used for controlling the energy of the dimension-reducing matrix $\Q_i$, which also has an effect of improving the conditioning of the subproblem w.r.t. $\Q_i$.
$h_i(\Q_i)=\mu_i\|\Q_i\|_{2,1}$ that we are mainly interested in has the ability of promoting rows of $\Q_i$ to be zeros (or approximately zeros), and thus can suppress the impact of the corresponding columns (features) in $\X_i$ -- which is effectively feature selection.
The function $h_i(\Q_i)=\mu_i\|\Q_i\|_{1,1}$ also does feature selection, but different columns of the dimension-reduced data, i.e., ${\bm X}_i{\Q_i}$, may use different features.
The regularizers in \eqref{eq:elastic1}-\eqref{eq:elastic2} are sometimes referred to as the \emph{elastic net regularizers} in statistics, which improve conditioning of the $\Q_i$-subproblem and perform feature selection at the same time.
$h_i(\Q_i)={\bm 1}_{+}({\bm Q}_i)$ is for restraining the canonical components to be non-negative so that ${\bm X}_i{\bm Q}_i$ maintains interpretability in some applications like video analysis -- where the columns of ${\bm X}_i{\bm Q}_i$ are weighted combinations of time frames \cite{fischer2007time}; joint nonnegativity and sparsity regularizers can also be considered \cite{sigg2007nonnegative}.
In this section, we propose an algorithm that can deal with the regularized and the original versions of MAX-VAR GCCA under a unified framework.

\subsection{Alternating Optimization}
To deal with Problem~\eqref{eq:CGCCA_reg}, our approach is founded on alternating optimization (AO); i.e.,
we solve two subproblems w.r.t. $\{\Q_i\}$ and $\G$, respectively.
As will be seen, such a simple strategy will lead to highly scalable algorithms in terms of both memory and computational cost.

To begin with, let us assume that after $r$ iterations the current iterate is $({\bm Q}^{(r)},\G^{(r)})$ where $\Q=[\Q_1^T,\ldots,\Q_I^T]^T$ and consider the subproblem
\begin{equation}\label{eq:Qi}
\min_{{\bm Q}_i}~(\nicefrac{1}{2})\left\|{\bm X}_i{\bm Q}_i-{\bm G}^{(r)}\right\|_F^2+ h_i(\Q_i),~\forall i.
\end{equation}
The above problem is a regularized least squares problem. When $\X_i$ is large and sparse,
many efficient algorithms can be considered to solve it.
For example, the alternating direction method of multipliers (ADMM) \cite{Boyd11} is frequently employed to handle Problem~\eqref{eq:Qi} in a scalable manner.
However, ADMM is a primal-dual method that does not guarantee monotonic decrease of the objective value, which will prove useful in later convergence analysis.
Hence, we propose to employ the proximal gradient (PG) method for handling Problem~\eqref{eq:Qi}.
To explain, let us denote $\Q_i^{(r,t)}$ as the $t$th update of $\Q_i$ when $\G^{(r)}$ is fixed.
Under this notation, we have $\Q_i^{(r,0)}=\Q_i^{(r)}$ and $\Q_i^{(r,T)}=\Q_i^{(r+1)}$.
Let us rewrite \eqref{eq:Qi} as
\begin{equation}
\min_{\Q_i}~f_i\left(\Q_i,\G^{(r)}\right) + g_i(\Q_i),
\end{equation}
where we define $f_i(\Q_i,\G^{(r)})$ and $g_i(\Q_i)$ as the continuously differentiable part and the non-smooth part of the objective function in \eqref{eq:Qi}, respectively.
We also define $\nabla_{\Q_i} f_i(\Q_i,\G_i^{(r)})$ as the partial derivative of the differentiable part w.r.t. $\Q_i$.
When ``single-component'' regularizers such as $h_i(\Q_i)=\|\Q_i\|_{2,1}$ are employed, we have $f_i\left(\Q_i,\G^{(r)}\right)=(\nicefrac{1}{2})\left\|{\bm X}_i{\bm Q}_i-{\bm G}^{(r)}\right\|_F^2$ and $g_i(\Q_i)=h_i(\Q_i)$;
when $h_i(\Q_i)$ has multiple components such as $h_i(\Q_i)=\nicefrac{\mu_i}{2}\cdot\|\Q_i\|_{F}^2 + \beta_i \cdot\|\Q_i\|_{1,1}$, we have $f_i\left(\Q_i,\G^{(r)}\right)=(\nicefrac{1}{2})\left\|{\bm X}_i{\bm Q}_i-{\bm G}^{(r)}\right\|_F^2+\nicefrac{\mu_i}{2}\|\Q_i\|_F^2$ and $g_i(\Q_i)=\beta_i \cdot\|\Q_i\|_{1,1}$.

Per PG, we update $\Q_i$ by the following rule:
\begin{align}\label{eq:prox_gradient_Q}
        {\bm Q}_i^{(r,t+1)} &\leftarrow \texttt{prox}_{\alpha_ig_i}\left({\Q}_i^{(r,t)}-\alpha_i \nabla_{\Q_i} f_i\left({\Q}_i^{(r,t)},\G_i^{(r)}\right)\right) \\
       &= \arg\min_{\Q_i}~\frac{1}{2}\left\|{\Q}_i- {\bm H}_i^{(r,t)} \right\|_F^2+g_i({\bm Q}_i)\nonumber
\end{align}
where ${\bm H}_i^{(r,t)} ={\Q}_i^{(r,t)}-\alpha_i \nabla_{\Q_i} f_i({\Q}_i^{(r,t)},\G_i^{(r)})$.
For many $g_i(\cdot)$'s, the proximity operator in \eqref{eq:prox_gradient_Q} has closed-form or lightweight solutions \cite{parikh2013proximal}.

For example, if one adopts $g_i(\Q_i)=\mu_i\|\Q_i\|_{2,1}$, the update rule becomes
\begin{equation*}\label{eq:L21}
\Q_i^{(r,t)}(m,:) \leftarrow \begin{cases}
\bm 0, \quad \quad\quad\|{\bm H}_i^{(r,t)}(m,:)\|_2 < \mu_i,\\
\left(1 - \frac{\mu_i}{\|{\bm H}_i^{(r,t)}(m,:)\|_2}\right)\|{\bm H}_i^{(r,t)}(m,:)\|_2,~{\rm o.w.}
\end{cases}
\end{equation*}
For $g_i(\Q_i)=\mu_i\|\Q_i\|_{1,1}$, the update rule is similar to the above,
which is known as the \emph{soft-thresholding operator}.
For $g_i(\Q_i)={\bm 1}_{+}({\Q}_i)$, the solution is simply $\Q_i^{(r,t)}=\max\{{\bm H}_i^{(r,t)},{\bm 0}\}$.
An even simpler case is $h_i(\Q_i)=(\mu_i/2)\|\Q_i\|_{F}^2$; for this case, the update of $\Q_i$ is simply gradient descent, i.e.,
\begin{equation*}
\Q_i^{(r,t+1)} \leftarrow  \Q_i^{(r,t)} - \alpha_i\left(({\bm X}_i^T{\bm X}_i+\mu_i\bm I){\bm Q}_i^{(r,t)}- {\bm X}_i^T{\bm G}^{(r)}\right),
\end{equation*}
since the $\Q_i$-subproblem in \eqref{eq:Qi} does not have a non-smooth part.

By updating $\Q_i$ using the rule in \eqref{eq:prox_gradient_Q} for $T$ times where $T\geq 1$, we obtain
$\Q_i^{(r+1)}$.
Next, we consider solving the subproblem w.r.t. ${\bm G}$ when fixing $\{{\bm Q}_i\}_{i=1}^I$.
The $\G$-subproblem amounts to solving the following:
\begin{align}\label{eq:G-sub}
\min_{{\bm G}^T{\bm G}={\bm I}}~\sum_{i=1}^I\nicefrac{1}{2}\left\|{\bm X}_i{\bm Q}_i^{(r+1)}-{\bm G}\right\|_F^2.
\end{align}
Expanding the above and dropping the constants, we come up with the following equivalent problem:
\begin{align*}
&\max_{{\bm G}^T{\bm G}={\bm I}}~{\rm Tr}\left({\bm G}^T\sum_{i=1}^I{\bm X}_i{\bm Q}_i^{(r+1)}/I\right). \label{eq:G_sln}
\end{align*}
An optimal solution of ${\bm G}$ is the so-called Procrustes projection \cite{schonemann1966generalized}, which is implemented as follows:
Let ${\bm R}=\sum_{i=1}^I{\bm X}_i{\bm Q}_i^{(r+1)}.$
Then, we have
\[{\bm G}^{(r+1)} \leftarrow {\bm U}_R{\bm V}^T_R,\]
where ${\bm U}_R{\bm \Sigma}_R{\bm V}^T_R = {\rm svd}\left({\bm R},{\rm 'econ'}\right)$,
and ${\rm svd}\left(\cdot,{\rm 'econ'}\right)$ denotes the economy-size SVD that produces ${\bm U}_R\in\mathbb{R}^{L\times K}$,
${\bm \Sigma}_R\in\mathbb{R}^{K\times K}$ and ${\bm V}_R^T\in\mathbb{R}^{K\times K}$.
The above update is optimal in terms of solving the subproblem.
However, since this subproblem has multiple optimal solutions, picking an arbitrary one from the solution set results in difficulties in analyzing some aspects of the algorithm (specifically, the rate of convergence).
To overcome this issue, we propose to solve the following
\begin{equation}\label{eq:G-sub-prox}
\min_{\G^T\G={\bm I}}~\sum_{i=1}^I\frac{1}{2}\left\|{\bm X}_i{\bm Q}_i^{(r+1)}-{\bm G}\right\|_F^2+\omega\cdot\left\|{\bm G}-{\bm G}^{(r)}\right\|_F^2,
\end{equation}
where $\omega = \nicefrac{(1-\gamma)I}{2\gamma}$ and $\gamma\in(0,1]$. Note that $\omega\geq 0$ and the proximal term is added to ensure that $\G^{(r+1)}$ will not wander very far from $\G^{(r)}$.
An optimal solution to the above is still simple: The only change to the original $\G$-solution
is to use the following modified ${\bm R}$
\begin{equation}
 {\bm R}=\gamma\sum_{i=1}^I{\bm X}_i{\bm Q}_i^{(r+1)}/I + (1-\gamma)\G^{(r)},
\end{equation}
and the other operations (e.g., the economy-size SVD) remain the same.
While the addition of the proximal term may seem to ``degrade'' an optimal solution of the $\G$-subproblem (i.e., Problem~\eqref{eq:G-sub}) to an inexact one, this simple change helps establish nice convergence rate properties of the overall algorithm, as we will see.

\color{black}

The algorithm is summarized in Algorithm~\ref{algo:AltCCA}, which we call the alternating optimization-based MAX-VAR GCCA (\texttt{AltMaxVar}).
As one can see, the algorithm does not instantiate any large dense matrix during the procedure and thus is highly efficient in terms of memory.
Also, the procedure does not destroy sparsity of the data, and thus the computational burden is light when the data is sparse -- which is often the case in large-scale learning applications.
Detailed complexity analysis will be presented in the next subsection.

\begin{algorithm}

	{\small
	\SetKwInOut{Input}{input}
	\SetKwInOut{Output}{output}
	\SetKwRepeat{Repeat}{repeat}{until}
	
	\Input{$\{{\bm X}_i,\mu_i,\alpha_i\}_{i=1}^I$; $\gamma\in(0,1]$; $K$; $T$; $(\{{\bm Q}_i^{(0)}\}_{i=1}^I,{\bm G}^{(0)})$. }

	$r \leftarrow  0$;
	
	\Repeat{Some stopping criterion is reached}{
		$t\leftarrow 0$;
		
		${\bm E}_i^{(t)} \leftarrow {\bm Q}_i^{(r)}$ for $i=1,\ldots,I$;
		
       \While{$t\leq T$ and convergence not reached}{
       for all $i$, update

        ${\bm H}_i^{(r,t)} \leftarrow \Q_i^{(r,t)}-\alpha_i\nabla_{\Q_i} f_i\left(\Q_i^{(r,t)};\G_i^{(r)}\right)$;

		${\bm Q}_i^{(r,t+1)}\leftarrow \texttt{prox}_{\alpha_ig_i}\left( {\bm H}_i^{(r,t)} \right)$;
		
		$t \leftarrow t + 1$;
		}
		
		${\bm Q}_i^{(r+1)}\leftarrow{\bm Q}_i^{(r,T)}$;
		
		${\bm R}\leftarrow \gamma{\sum_{i=1}^I{\bm X}_i{\bm Q}_i^{(r+1)}}/I + (1-\gamma)\G^{(r)}$;
		
		${\bm U}_R{\bm \Sigma}_R{\bm V}_R^T\leftarrow {\rm svd}\left({\bm R},{\rm '{\rm econ}'}\right)$;
		
		${\bm G}^{(r+1)}\leftarrow{\bm U}_R{\bm V}_R^T$;
		
		$r \leftarrow r+1$;
	}
	
	\Output{$\left\{{\bm Q}_i^{(r)}\right\}_{i=1}^I$, ${\bm G}^{(r)}$}}
	\caption{\texttt{AltMaxVar}}\label{algo:AltCCA}

\end{algorithm}

\subsection{Computational and Memory Complexities}
The update rule in \eqref{eq:prox_gradient_Q} inherits the good features from
the PG method.
First, there is no ``heavy computation'' if the views $\X_i$ for $i=1,\ldots,I$ are sparse. Specifically, the major computation in the update rule of \eqref{eq:prox_gradient_Q} is computing the partial gradient of the smooth part of the cost function, i.e., $\nabla_{\Q_i} f_i(\Q_i,\G_i)$.
To this end, ${\bm X}_i{\bm Q}_i$ should be calculated first, since if ${\bm X}_i$ is sparse,
this matrix multiplication step has a complexity order of ${\cal O}({\rm nnz}({\bm X}_i)\cdot K)$ flops,
where ${\rm nnz}(\cdot)$ counts the number of non-zeros. The next multiplication, i.e., ${\bm X}_i^T({\bm X}_i{\bm Q}_i)$, has the same complexity order.
Similarly, the operation of $\X_i^T\G$ has the same complexity.
For solving the $\G$-subproblem, the major operation is the SVD of ${\bm R}$.
This step is also not computationally heavy -- what we ask for is an economy-size SVD of a very thin matrix (of size $L\times K$, $L\gg K$).
This has a complexity order of ${\cal O}(LK^2)$ flops \cite{GHGolub1996}, which is light.

In terms of memory, all the terms involved (i.e., ${\bm Q}_i$, ${\bm G}_i$, ${\bm X}_i{\bm Q}_i$, ${\bm X}_i^T{\bm X}_i{\bm Q}_i$ and ${\bm X}_i^T{\bm G}_i$) only require ${\cal O}(LK)$ memory or less,
but the eigen-decomposition-based solution needs ${\cal O}(M_i^2)$ and ${\cal O}(L^2)$ memory to store $({\bm X}_i^T{\bm X}_i)^{-1}$ and ${\bm M}$, respectively. Note that $K$ is usually very small (and up to our control) compared to $L$ and $M_i$,  which are approximately of the same large size in applications like word embedding.

\section{Convergence Properties}

In this section, we study convergence properties of \texttt{AltMaxVar}.
Note that the algorithm alternates between a (possibly) non-smooth subproblem and a manifold-constrained subproblem, and the subproblems may or may not be solved to optimality.
Existing convergence analysis for exact and inexact block coordinate descent such as those in \cite{bertsekas1999nonlinear,razaviyayn2013unified,xu2013block,xu2014globally} can not be directly applied to analyze \texttt{AltMaxVar}, and thus its convergence properties are not obvious.
For clarity of exposition, we first define a critical point, or, a KKT point, of Problem~\eqref{eq:CGCCA_reg}.
A KKT point $(\G^\ast,\Q^\ast)$ satisfies the following first-order optimality conditions:
\begin{align*}\label{eq:KKT}
\begin{cases}
 {\bm 0}\in \nabla_{\Q_i}~f_i(\Q_i^\ast,\G^\ast) + \partial_{\Q_i} g_i(\Q^\ast),~\forall i\\
 {\bm 0}= \nabla_{\G}~\sum_{i=1}^I f_i(\Q_i^\ast,\G^\ast) + \G^\ast{\bm \Lambda}^\ast,\quad (\G^\ast)^T\G^\ast = \bm I,
\end{cases}
\end{align*}
where ${\bm \Lambda}$
is a Lagrangian multiplier associated with the constraint $\G^T\G={\bm I}$, and
$\partial_{\Q_i} g_i(\Q_i)$ denotes a subgradient of the (possibly) non-smooth function $g_i(\Q_i)$.
We first show that
\begin{Prop}\label{lem:monotonicity}
Assume that $\alpha_i \leq 1/L_i$ for all $i$, where $L_i=\lambda_{\max}({\bm X}_i^T{\bm X}_i)$ is the largest eigenvalue of ${\bm X}_i^T{\bm X}_i$.
Also assume that $g_i(\cdot)$ is a closed convex function, $T\geq 1$, and $\gamma\in(0,1]$. Then, the following holds:
\begin{itemize}
\item[(a)] The objective value of Problem~\eqref{eq:CGCCA} is non-increasing.
In addition, every limit point of the solution sequence $\{{\bm G}^{(r)},\{{\bm Q}_i^{(r)}\}\}_{r}$ is a KKT point of Problem~\eqref{eq:CGCCA}.
\item[(b)] If ${\bm X}_i$ and ${\bm Q}^{(0)}_i$ for $i=1,\ldots,I$ are bounded and ${\rm rank}({\bm X}_i)=M_i$,
then, the whole solution sequence converges to the set ${\cal K}$ that consists of all the KKT points. 
\end{itemize}
\end{Prop}
Proposition~\ref{lem:monotonicity} (a) characterizes the limit points of the solution sequence: Even if only one proximal gradient step is performed in each iteration $r$, every \emph{convergent subsequence} of the solution sequence attains a KKT point of Problem~\eqref{eq:CGCCA_reg}.
As we demonstrate in the proof (relegated to the Appendix),
\texttt{AltMaxVar} can be viewed as an algorithm that successively deals with local upper bounds of
the two subproblems, which has a similar flavor as \emph{block successive upper bound minimization} (BSUM) \cite{razaviyayn2013unified}.
However, the generic BSUM framework does not cover nonconvex constraints such as $\G^T\G={\bm I}$,
Hence, the convergence properties of BSUM cannot be applied to show Proposition~\ref{lem:monotonicity}. To fill this gap, 
careful custom convergence analysis is provided in the appendix.
The (b) part of Proposition~\ref{lem:monotonicity} establishes the convergence of the whole solution sequence -- which is a much stronger result. The assumption ${\rm rank}(\X_i)=M_i$, on the other hand, is also relatively more restrictive.

It is also meaningful to estimate the number of iterations that is needed for the algorithm to reach a neighborhood of a KKT point.
To this end, let us define the following potential function:
\begin{align*}
&Z^{(r+1)}=\sum_{t=0}^{T-1}\sum_{i=1}^I\left\|\tilde{\nabla}_{\Q_i} F_i\left(\Q_i^{(r,t)},\G^{(r)}\right)\right\|_F^2\\
                                                   &+ \left\| \G^{(r)} - \nicefrac{\sum_{i=1}^I{\bm X}_i{\Q_i^{(r+1)}}}{I} + \G^{(r+1)}{\bm \Lambda}^{(r+1)} \right\|_F^2,
\end{align*}
\color{black}
where $F_i(\Q_i,\G)=f_i(\Q_i,\G)+g_i(\Q_i,\G)$,
${\bm \Lambda}^{(r+1)}$ is the Lagrangian multiplier associated with the solution ${\bm G}^{(r+1)}$, and

\begin{align*}
\tilde{\nabla}_{\Q_i} F_i(\Q_i^{(r,t)},\G^{(r)})=\frac{1}{\alpha_i}\left(\Q_i^{(r,t)} - \texttt{prox}_{\alpha_ig_i}\left({\bm H}_i^{(r,t)}\right)\right).
\end{align*}
Note that the update w.r.t. $\Q_i$ can be written as
$\Q_i^{(r,t+1)}={\Q}_i^{(r,t)}-\alpha_i \tilde{\nabla}_{\Q_i}F_i({\Q}_i^{(r,t)},\G_i^{(r)})$ \cite{parikh2013proximal}
-- and therefore $\tilde{\nabla}_{\Q_i} F_i(\Q_i^{(r,t)},\G^{(r)})$ is also called the \emph{proximal gradient} of the $\Q_i$-subproblem w.r.t. $\Q_i$ at $(\Q_i^{(r,t)},\G^{(r)})$, as a counterpart of the classic gradient that is defined on smooth functions.
One can see that $Z^{(r+1)}$ is a value that is determined by two consecutive outer iterates indexed by $r$ and $r+1$ of the algorithm.
$Z^{(r+1)}$ has the following property:
\begin{Lemma}\label{lem:z}
$Z^{(r+1)}\rightarrow 0$ implies that $\left(\{\Q_i^{(r)}\}_i,\G^{(r)}\right)$ approaches a KKT point.
\end{Lemma}
The proof of Lemma~\ref{lem:z} is in Appendix~\ref{app:lemma_z}.
As a result, we can use the value of $Z^{(r+1)}$
to measure how close is the current iterate to a KKT point, thereby estimating the iteration complexity.
Following this rationale, we show that

\begin{Theorem}\label{thm:complexity}
Assume that $\alpha_i<1/L_i$, $0<\gamma<1$ and $T\geq 1$.
Let $\delta> 0$ and $J$ be the number of iterations when $Z^{(r+1)}\leq \delta$ holds for the first time.
Then, there exists a constant $v$ such that
$\delta \leq \nicefrac{v}{J-1}$;
that is, the algorithm converges to a KKT point at least sublinearly.
\end{Theorem}
The proof of Theorem~\ref{thm:complexity} is relegated to Appendix~\ref{app:complexity}.
By Theorem~\ref{thm:complexity}, \texttt{AltMaxVar} reduces the optimality gap (measured by the $Z$-function) between the current iterate and a KKT point to ${\cal O}(1/r)$ after $r$ iterations.
One subtle point that is worth mentioning is that the analysis in Theorem~\ref{thm:complexity} holds when $\gamma<1$ -- which corresponds to the case where the ${\bm G}$-subproblem in \eqref{eq:G-sub} is \emph{not} optimally solved (to be specific, what we solve is a local surrogate in \eqref{eq:G-sub-prox}). This reflects some interesting facts in AO -- when the subproblems are handled in a more conservative way using a controlled step size, convergence rate may be guaranteed.
On the other hand, more conservative step sizes may result in slower convergence. Hence, choosing an optimization strategy usually poses a trade-off between practical considerations such as speed and theoretical guarantees.

Proposition~\ref{lem:monotonicity} and Theorem~\ref{thm:complexity} characterize convergence properties of \texttt{AltMaxVar} with a general regularization term $h_i(\cdot)$.
It is also interesting to consider the special case where $h_i(\cdot)=(\mu_i/2)\|\cdot\|_F^2$ -- which correspond to the original MAX-VAR formulation (when $\mu_i=0$) and its ``diagonally loaded'' version ($\mu_i>0$).
The corresponding problem is \emph{optimally solvable} via taking the $K$ leading eigenvectors of ${\bm M}=\sum_{i=1}^{I}{\bm X}_i({\bm X}_i^T{\bm X}_i+\mu_i{\bm I})^{-1}{\bm X}_i^T$ \cite{rastogimultiview}.
It is natural to wonder if \texttt{AltMaxVar} has sacrificed optimality in dealing with this special case for the sake of gaining scalability? The answer is -- thankfully -- not really. This is not entirely surprising; to explain, let us denote ${\bm U}_1= {\bm U}_M(:,1:K)$ and ${\bm U}_2={\bm U}_M(:,K+1:L)$ as the $K$ principal eigenvectors of ${\bm M}$ and the eigenvectors spanning its orthogonal complement, respectively.
Recall that our ultimate goal is to find ${\bm G}$ that is a basis of the range space of ${\bm U}_1$, denoted by ${\cal R}({\bm U}_1)$.
Hence, the speed of convergence can be measured through the distance between ${\cal R}({\bm G})$ and
${\cal R}({\bm U}_1)$.
To this end, we adopt
the definition of subspace distance in \cite{GHGolub1996}, i.e.,
${\rm dist}\left({\cal R}({\bm G}^{(r)}),{\cal R}({\bm U}_1)\right) =\|{\bm U}_2^T{\bm G}^{(r)}\|_2$ and show that
\begin{Theorem} \label{thm:main}
   Denote the eigenvalues of ${\bm M}\in\mathbb{R}^{L\times L}$ by $\lambda_1,\ldots,\lambda_L$ in descending order.
   Consider $h_i(\cdot)=\frac{\mu_i}{2}\|\cdot\|_F^2$ for $\mu_i\geq 0$ and let $\gamma = 1$.
   Assume that ${\rm rank}({\bm X}_i)=M_i$, $\lambda_K>\lambda_{K+1}$, and ${\cal R}({\bm G}^{(0)})$ is not orthogonal to any component in ${\cal R}({\bm U}_1)$, i.e., 
   \begin{equation}\label{eq:sigma}
   \begin{aligned}
    \cos(\theta)&=\min_{{\bm u}\in{\cal R}({\bm U}_1), {\bm v}\in{\cal R}({\bm G}^{(0)})}\frac{|{\bm u}^T{\bm v}|}{(\|{\bm u}\|_2\|{\bm v}\|_2)}\\
                     &=\sigma_{\min}({\bm U}_1^T{\bm G}^{(0)})>0.
   \end{aligned}
   \end{equation} 
   In addition, assume that each subproblem in \eqref{eq:Qi} is solved to accuracy $\epsilon^{(r)}$ at iteration $r$, i.e.,
   $\|\Q_i^{(r)}-\tilde{\Q}_i^{(r)}\|_2 \leq \epsilon^{(r)}$, where $\tilde{\Q}_i^{(r)}=(\X_i^T\X_i+\mu_i{\bm I})^{-1}\X_i^T\G^{(r-1)}$.
   Assume that $\epsilon^{(r)}$ is sufficiently small, i.e.,
  \begin{align}\label{eq:converge_suff}
\epsilon^{(r)} &\leq \frac{\lambda_K - \lambda_{K+1}}{3\sum_{i=1}^I\lambda_{\max}(\X_i)} \\
&\times \min\left\{\sigma_{\min}\left( \U_2^T\G^{(r)} \right),\sigma_{\max}\left( \U_1^T\G^{(r)} \right)\right\}.  \nonumber 
   \end{align} 
   Then, ${\rm dist}\left({\cal R}({\bm G}^{(r)}),{\cal R}({\bm U}_1)\right)$ approaches zero at a linear rate; i.e.,
   \[   {\rm dist}\left({\cal R}({\bm G}^{(r)}),{\cal R}({\bm U}_1)\right) \leq    \left( \frac{2\lambda_{K+1}+\lambda_K}{2\lambda_K+\lambda_{K+1}} \right)^{r} \tan(\theta).  \]
\end{Theorem}
Theorem~\ref{thm:main} ensures that if a $T$ suffices for the $\Q$-subproblem to obtain a good enough approximation of the solution of Problem~\eqref{eq:Qi}, the algorithm converges \textit{linearly} to a \emph{global optimal} solution -- this means that we have gained scalability using \texttt{AltMaxVar}
without losing optimality.
Note that \eqref{eq:converge_suff} means that the $\Q_i$-subproblem may require a higher solution accuracy when ${\cal R}(\G^{(r)})$ approaches ${\cal R}(\U_1)$, since $\sigma_{\min}(\U_2^T\G^{(r)})$ is close to zero under such circumstances.
Nevertheless, since the result is based on worst-case analysis, the solution of the $\Q_i$-subproblem can be far rougher in practice -- and one can still observe good convergence behavior of \texttt{AltMaxVar}. 
In fact, in our simulations, we observe that using $T=1$ already gives very satisfactory results (as will be shown in the next section), which leads to computationally very cheap updates.


\begin{Remark}
According to the proof of Theorem~\ref{thm:main},
when dealing with Problem~\eqref{eq:CGCCA_reg} with $h_i(\cdot)=\nicefrac{\mu_i}{2}\|\cdot\|_F^2$, the procedure of \texttt{AltMaxVar} can be interpreted as a variant of the orthogonal iteration \cite{GHGolub1996}.
Based on this insight, many other approaches can be taken; e.g., the $\Q_i$-subproblem can be handled by conjugate gradient and the SVD step can be replaced by the QR decomposition -- which may lead to computationally even cheaper updates.
Nevertheless, our interest lies in solving \eqref{eq:CGCCA_reg} with a variety of regularizations under a unified framework, and the aforementioned alternatives cannot easily handle other regularizations.
\end{Remark}

\vspace{-.2cm}

\section{Numerical Results}
\vspace{-.1cm}
In this section, we use synthetic data and real experiments to showcase the effectiveness of the proposed algorithm.
Throughout this section, the step size of the $\Q_i$-subproblem of \texttt{AltMaxVar} is set to be $\alpha_i=0.99\times 1/\lambda_{\max}(\X_i^T\X_i)$. When $h_i(\Q_i)=\nicefrac{\mu_i}{2}\|\Q_i\|_F^2$ is employed, we let $\gamma = 1$ following Theorem~\ref{thm:main}; otherwise, we let $\gamma = 0.9999$ -- so that the convergence rate guarantee in Theorem~\ref{thm:complexity} holds.
All the experiments are coded in \texttt{Matlab} and conducted on a Linux server equipped with 32 1.2GHz cores and 128GB RAM.

\vspace{-.45cm}
\subsection{Sanity Check: Small-Size Problems}
We first use small-size problem instances to verify the convergence properties
that were discussed in the last section.
\subsubsection{Classic MAX-VAR GCCA}
We generate the synthetic data in the following way: First, we let ${\bm Z}\in\mathbb{R}^{L\times N}$ be a common latent factor of different views,
where the entries of ${\bm Z}$ are drawn from the zero-mean i.i.d. Gaussian distribution and $L\geq N$.
Then, a `mixing matrix' ${\bm A}_i\in\mathbb{R}^{N\times M_i}$ is multiplied to ${\bm Z}$, resulting in ${\bm Y}_i={\bm Z}{\bm A}_i$.
We let $M_1=\ldots=M_I=M$ in this section.
Finally, we add noise so that ${\bm X}_i={\bm Y}_i + \sigma{\bm N}_i$.
Here, ${\bm A}_i$ and ${\bm N}_i$ are generated in the same way as ${\bm Z}$.
We first apply the algorithm with the regularization term $h_i(\cdot)=\nicefrac{\mu_i}{2}\|\cdot\|_F^2$ and let $\mu_i=0.1$.
Since $L$ and $M$ are small in this subsection, we employ the optimal solution that is based on eigen-decomposition as a baseline.
The multiview latent semantic analysis (\texttt{MVLSA}) algorithm that was proposed in \cite{rastogimultiview} is also employed as a baseline.
In this section, we stop \texttt{AltMaxVar} when the absolute change of the objective value is smaller than $10^{-4}$.
\begin{figure}[ht]
\centering
\includegraphics[width=.55\linewidth]{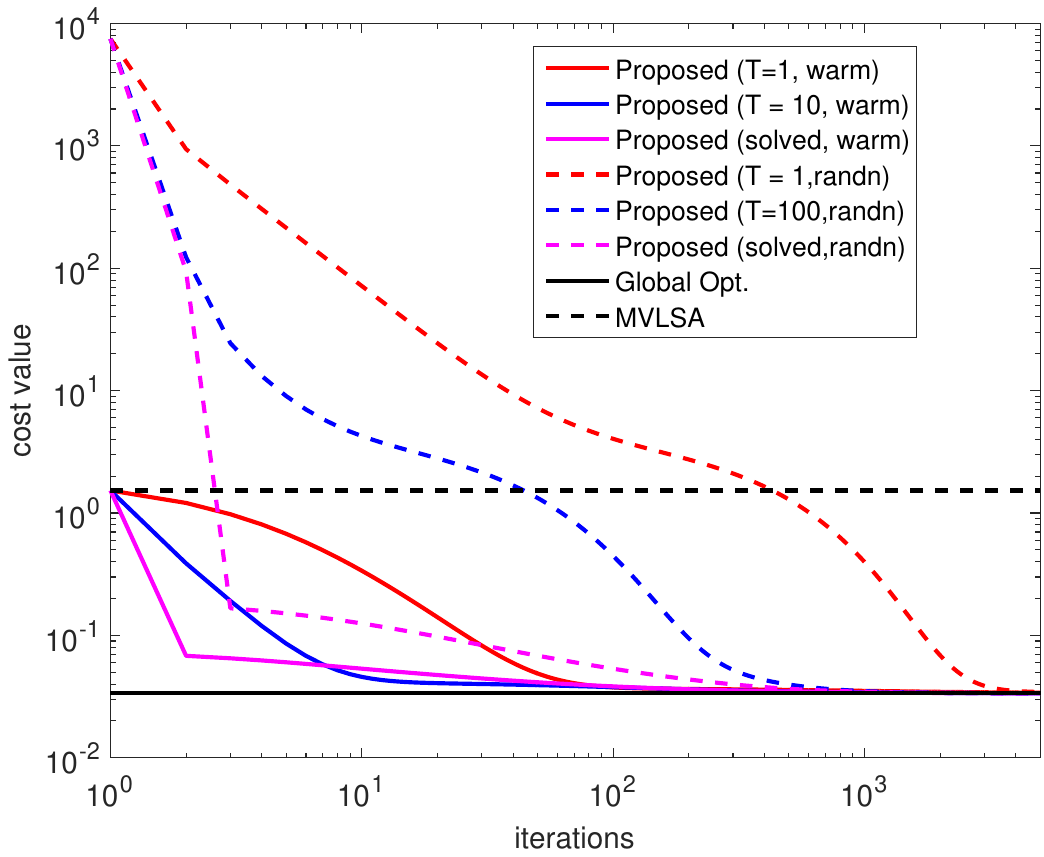}
\caption{Convergence curves of the algorithms.}
\label{fig:subfigure1}
\end{figure}

In Fig.~\ref{fig:subfigure1}, we let $(L,M,N,I)=(500,25,20,3)$.
We set $\sigma=0.1$ in this case, let $P=8$ and $\gamma=1$ for \texttt{MVLSA} and \texttt{AltMaxVar}, respectively, and ask for $K=5$ canonical components.
The results are averaged over 50 random trials, where ${\bm Z}$, $\{{\bm A}_i\}$, $\{{\bm N}_i\}$ are randomly generated in each trial.
We test the proposed algorithm under different settings:
We let $T=1$, $T=10$, and the gradient descent run until the inner loop converges (denoted as `solved' in the figures).
We also initialize the algorithm with random initializations (denoted as `randn') and warm starts (denoted as `warm') -- i.e., using the solutions of \texttt{MVLSA} as starting points.
Some observations from Fig.~\ref{fig:subfigure1} are in order.
First, the proposed algorithm using various $T$'s including $T=1$ and random initialization can reach the global optimum, which supports the analysis in Theorem~\ref{thm:main}.
Second, by increasing $T$, the overall cost value decreases faster in terms of number of outer iterations -- using $T=10$ already gives very good speed of decreasing the cost value.
Third, \texttt{MVLSA} cannot attain the global optimum, as expected. However, it provides good initialization: Using the warm start, the cost value comes close to the optimal value within 100 iterations in this case, even when $T=1$ is employed.
In fact, the combination of \texttt{MVLSA}-based initialization and using $T=1$ offers the most computationally efficient way of implementating the proposed algorithm -- especially for the large-scale case.
In the remaining part of this section, we will employ \texttt{MVLSA} as the initialization of \texttt{AltMaxVar} and employ $T=1$ for the $\Q$-subproblem.

\subsubsection{Feature-Selective MAX-VAR GCCA}
To test the proposed algorithm with non-smooth regularizers,
we generate cases where outlying features are present in all views.
Specifically, we let ${\bm X}_i=[{\bm Z}{\bm A}_i,{\bm O}_i] + \sigma{\bm N}_i$,
where ${\bm O}_i\in\mathbb{R}^{L\times N_o}$ denotes the irrelevant outlying features and the elements of ${\bm O}_i$ follow the i.i.d. zero-mean unit-variance Gaussian distribution.
We wish to perform MAX-VAR GCCA of the views while discounting ${\bm O}_i$ at the same time.
To deal with outlying features, we employ the regularizer $g_i(\cdot)=\mu_i\|\cdot\|_{2,1}$ and implement the algorithm with $\mu_i=0.5$ and $\mu_i=1$, respectively.
Under this setting, the optimal solution to Problem \eqref{eq:CGCCA_reg} is unknown.
Therefore, we evaluate the performance by observing
$
{\rm metric}_1 =\nicefrac{1}{I}\sum_{i=1}^I\|{\bm X}_i(:,{\cal S}_i^c)\hat{\Q}_i({\cal S}_i^c,:)-\hat{\bm G}\|_F^2$
and  ${\rm metric}_2 = \nicefrac{1}{I}\sum_{i=1}^I\|{\bm X}_i(:,{\cal S}_i)\hat{\Q}_i({\cal S}_i,:)\|_F^2$,
where ${\cal S}^c_i$ and ${\cal S}_i$ denote the index sets of ``clean'' and outlying features of view $i$, respectively -- i.e., ${\bm X}_i(:,{\cal S}_i^c)={\bm Z}_i{\bm A}_i$ and ${\bm X}_i(:,{\cal S}_i)={\bm O}_i$ if noise is absent.
${\rm metric}_1$ measures the performance of matching $\hat{\bm G}$ with the relevant part of the views, while ${\rm metric}_2$ measures the performance of suppressing the irrelevant part.
We wish that our algorithm yields low values of ${\rm metric}_1$ and ${\rm metric}_2$ simultaneously.

\begin{figure}[ht]
\centering
{\includegraphics[width=.7\linewidth]{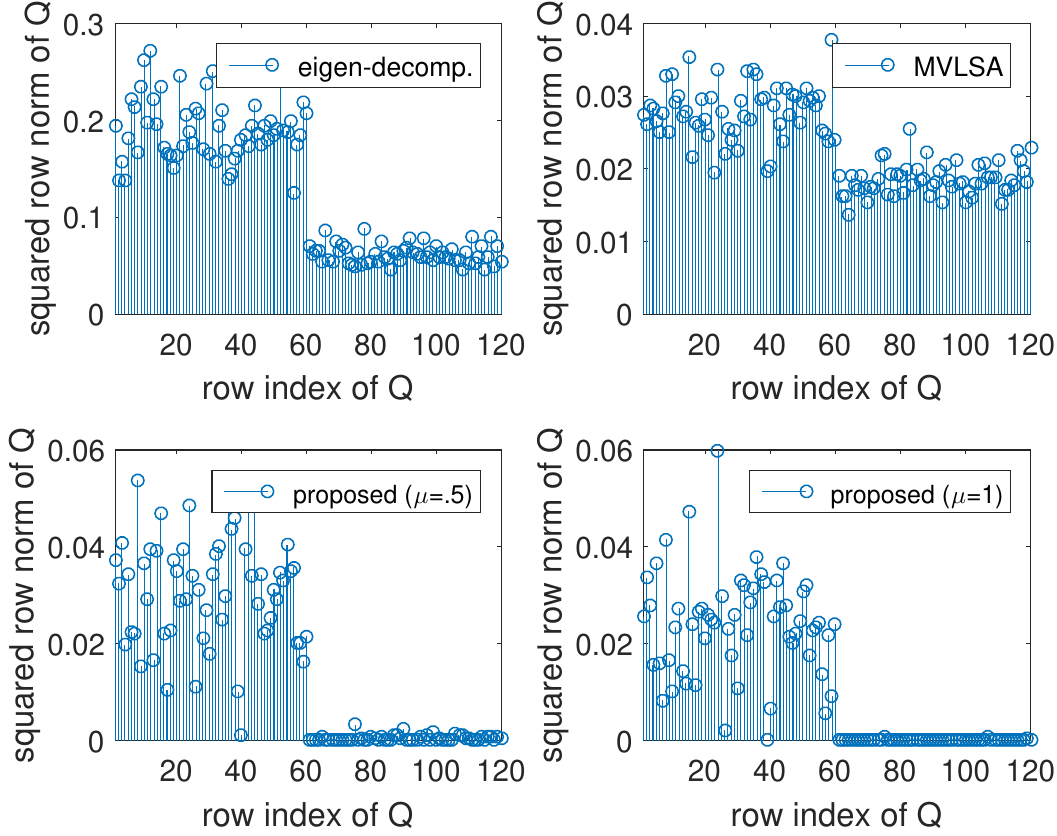}
\caption{Average row-norms of ${\bm Q}_i$ (i.e., $(1/I)\sum_{i=1}^I\|{\bm Q}_i(m,:)\|_2^2$) for all $m$ given by the algorithms.}
\label{fig:small_row_norm}}
\end{figure}

Table~\ref{tab:small_sparse} presents the results of a small-size case which are averaged from 50 random trials, where $(L,M,N,I)=(150,60,60,3)$ and $|{\cal S}|=\{61,\ldots,120\}$; i.e.,
$60$ out of $120$ features of $\X_i\in\mathbb{R}^{150\times 120}$ are outlying features.
The average power of the outlying features is set to be the same as that of the clean features, i.e.,
$\|{\bm Q}_i\|_F^2/L|{\cal S}_i|=\|{\bm Z}{\bm A}_i\|_F^2/LM$ so that the outlying features are not negligible.
We ask for $K=10$ canonical components. For \texttt{MVLSA}, we let the rank-truncation parameter to be $P=50$.
One can see that the eigen-decomposition based algorithm gives similar high values of both the evaluation metrics since it treats $\X_i(:,{\cal S}^c_i)$ and $\X_i(:,{\cal S}_i)$ equally.
It is interesting to see that \texttt{MVLSA} suppresses the irrelevant features to some extent -- although
it does not explicitly consider outlying features, our understanding is that the PCA pre-processing on the views can somewhat suppress the outliers.
Nevertheless, \texttt{MVLSA} does not fit the relevant part of the views well.
The proposed algorithm gives the lowest values of both metrics. In particular, when $\mu_i=1$ for all $i$, the irrelevant part is almost suppressed completely.
Another observation is that using $\mu_i=0.5$, the obtained score of ${\rm metric}_1$ is slightly lower than that under $\mu_i=1$, which makes sense since the algorithm pays more attention to feature selection using a larger $\mu$.
An illustrative example using a random trial can be seen in Fig.~\ref{fig:small_row_norm}.
From there, one can see that the proposed algorithm gives ${\bm Q}_i$'s with almost zero rows over ${\cal S}$, thereby performing feature selection.

\begin{table}[htbp]
  \centering
  \caption{Performance of the algorithms when irrelevant features are present. $(L,M,N)=(150,60,60)$; $|{\cal S}|=60$; $\X_i\in\mathbb{R}^{150\times 120}$; $\sigma=1$.}
   	\resizebox{.5\linewidth}{!}{
    \begin{tabular}{c|c|c}
    \hline
    \hline
    Algorithm & ${\rm metric}_1$ &  ${\rm metric}_2$ \\
    \hline
    \hline
    eigen-decomp &9.547&9.547 \\
    \hline
    \texttt{MVLSA} & 15.506	&1.456 \\
    \hline
    proposed ($\mu=.5$) & \textbf{0.486}&	$9.689\times 10^{-3}$ \\
    \hline
    proposed ($\mu=1$) & 1.074 &	${\bf 8.395\times 10^{-4}}$ \\
    \hline
    \hline
    \end{tabular}%
    }
  \label{tab:small_sparse}%
\end{table}%

\subsection{Scalability Test: Large-Size Problems}

\subsubsection{Original MAX-VAR GCCA}
We first test the case where no outlying features are involved
and the regularizer $h_i(\cdot)=\nicefrac{\mu_i}{2}\|\cdot\|_F^2$ is employed.
The views $\X_i={\bm Z}{\bm A}_i+\sigma {\bm N}_i$ are generated following a similar way as in the last subsection, but ${\bm Z}$, ${\bm A}_i$ and ${\bm N}_i$ are sparse so that ${\bm X}_i$ are sparse with a density level $\rho_i$ that is definied as $\rho_i = \frac{{\rm nnz}({\bm X}_i)}{LM}$.
In the simulations, we let $\rho=\rho_1=\ldots=\rho_I$.
In the large-scale cases in this subsection,
and the results are obtained via averaging 10 random trials.

\begin{figure}[ht]
\centering
{\includegraphics[width=.5\linewidth]{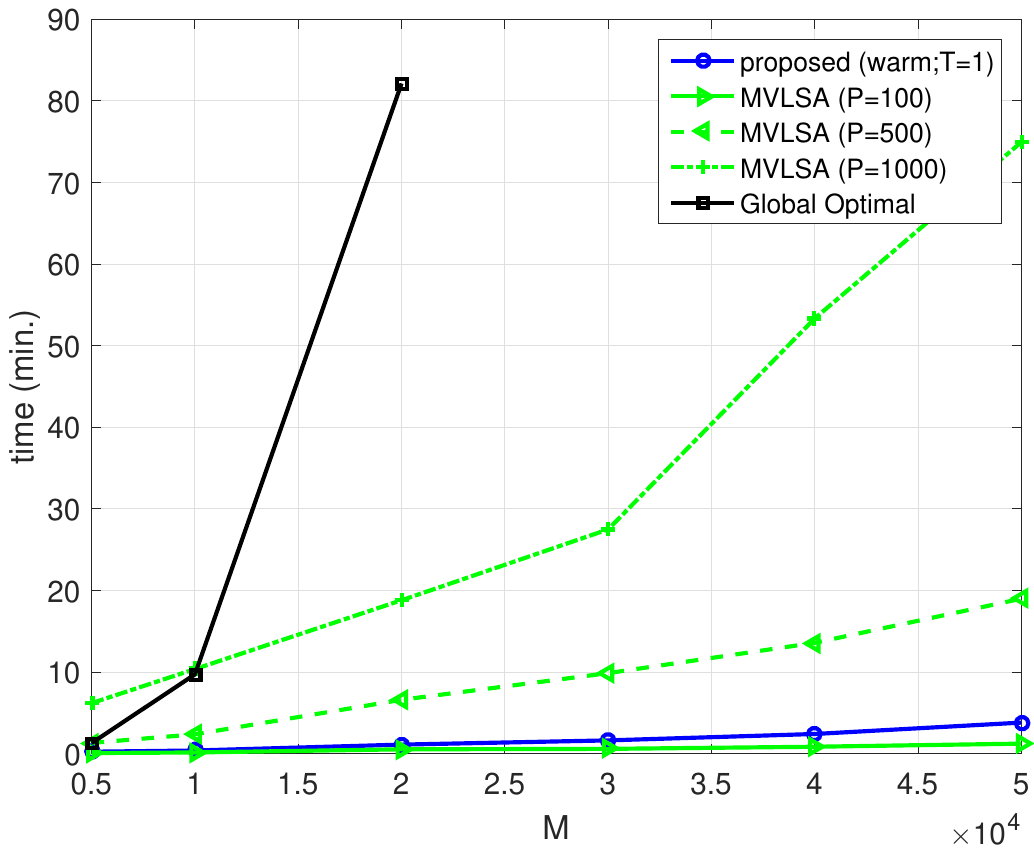}
\caption{Runtime of the algorithms for various problem sizes. $L=M/0.8$, $\rho\leq 10^{-3}$, $\sigma=0.1$.}
\label{fig:runtime_plain}}
\end{figure}

In Fig.~\ref{fig:runtime_plain}, we show the runtime performance of the algorithms for various sizes of the views,
where density of the views is controlled so that $\rho \approx 10^{-3}$.
The regularization parameter $\mu_i=0.1$ is employed by all algorithms.
We let $M=L\times 0.8$, $M=N$ and change $M$ from $5,000$ to $50,000$.
To run \texttt{MVLSA}, we truncate the ranks of views to $P=100$, $P=500$ and $P=1,000$, respectively.
We use \texttt{MVLSA} with $P=100$ to initialize \texttt{AltMaxVar} and let $T=1$ and $\gamma=1$. We stop the proposed algorithm when the absolute change of the objective value is smaller than $10^{-4}$.
Ten random trials are used to obtain the results.
One can see that the eigen-decomposition based algorithm does not scale well since the matrix $({\bm X}_i^T{\bm X}_i+\mu_i{\bm I})^{-1}$ is dense.
In particular, the algorithm exhausts the memory quota (32GB RAM) when $M=30,000$.
\texttt{MVLSA} with $P=100$ and the proposed algorithm both scale very well from $M=5,000$ to $M=50,000$:
When $M=20,000$, brute-force eigen-decomposition takes almost 80 minutes,
whereas \texttt{MVLSA} ($P=100$) and \texttt{AltMaxVar} both use less than 2 minutes.
Note that the runtime of the proposed algorithm already includes the runtime of the initialization time by \texttt{MVLSA} with $P=100$, and thus the runtime curve of \texttt{AltMaxVar} is slightly higher than that of \texttt{MVLSA} ($P=100$) in Fig.~\ref{fig:runtime_plain}.
Another observation is that, although \texttt{MVLSA} exhibits good runtime performance when using $P=100$,
its runtime under $P=500$ and $P=1,000$ is not very appealing.
The corresponding cost values can be seen in Table~\ref{tab:cost_large_plain}.
The eigen-decomposition based method gives the lowest cost values when applicable, as it is an optimal solution.
The proposed algorithm gives favorable cost values that are close to the optimal ones, even when only one iteration of the ${\bm Q}$-subproblem is implemented for every fixed $\G^{(r)}$ -- this result supports our analysis in Theorem~\ref{thm:main}.
Increasing $P$ helps improve \texttt{MVLSA}.
However, even when $P=1,000$, the cost value given by \texttt{MVLSA} is still higher than that of \texttt{AltMaxVar},
and \texttt{MVLSA} using $P=1,000$ is much slower than \texttt{AltMaxVar}.

\begin{table}[htbp]
  \centering
   \caption{Cost values of the algorithms for different problem sizes. $L=M/0.8$, $\rho= 10^{-3}$, $\sigma=0.1$. $\dagger$ means `out-of-memory'.}
    	\resizebox{.5\linewidth}{!}{\huge
    \begin{tabular}{c|c|c|c|c|c|c}
    \hline
    \hline
    \multirow{2}[4]{*}{Algorithm} & \multicolumn{6}{c}{$M$} \\
\cline{2-7}          & 5,000  & 10,000 & 20,000 & 30,000 & 40,000 & 50,000 \\
    \hline
    \hline
    Global Opt & 0.053 & 0.033 & 0.021 & $\dagger$& $\dagger$ & $\dagger$  \\
    \hline
    MVLSA ($P=100$) & 2.164 & 3.527 & 5.065 & 5.893 & 6.475 & 7.058 \\
    \hline
    MVLSA ($P=500$) & 0.280 & 0.717 & 1.766 & 2.582 & 3.407 & 3.996 \\
    \hline
    MVLSA ($P=1,000$) & 0.125 & 0.287 & 0.854 & 1.406 & 2.012 & 2.513 \\
    \hline
    Proposed  & \textbf{0.092} & \textbf{0.061} &\textbf{ 0.049 }& \textbf{0.043} & \textbf{0.038} & \textbf{0.039} \\
    \hline
    \hline
    \end{tabular}%
}%
  \label{tab:cost_large_plain}%
\end{table}%


\subsubsection{Feature-Selective MAX-VAR GCCA}
Table~\ref{tab:rhochange} presents the simulation results of a large-scale case
in the presence of outlying features.
Here, we fix $L=100,000$ and $M=80,000$ and change the density level $\rho$.
We add $|{\cal S}_i|=30,000$ outlying features to each view and every outlying feature is a random sparse vector whose non-zero elements follow the zero-mean i.i.d. unit-variance Gaussian distribution.
We also scale the outlying features as before so that the energy of the clean and outlying features are comparable.
The other settings follow those in the last simulation.
One can see from Table~\ref{tab:rhochange} that the proposed algorithm with $\mu_i=0.05$
gives the most balanced result -- both evaluation metrics in with fairly low levels.
Using $\mu_i=0.5$ suppresses the $\Q_i({\cal S},:)$ quite well, but using a larger $\mu_i$ also brings some sacrifice to the fitting metric.
In terms of runtime, one can see that the proposed algorithm operates within the same order of magnitude of time as \texttt{MVLSA} needs.
Note that the proposed algorithm works with the intact views of size $L\times M$, while \texttt{MVLSA} works with heavily truncated data.
Therefore, such runtime performance of \texttt{AltMaxVar} is very satisfactory.

Similar results can be seen in Table~\ref{tab:Ichange}, where we let $\rho = 10^{-4}$ and change $I$ from $3$ to $8$. One can see that increasing the number of views does not increase the runtime of the proposed algorithm. The reason is that the updates of different $\Q_i$'s can be easily parallelized since the subproblems w.r.t. ${\bm Q}_i$'s are separable.
One can implement the parallel computations using the \texttt{parfor} function of \texttt{Matlab}.

\begin{table}[htbp]
  \centering
  \caption{Evaluation of the algorithm for different data densities in the presence of outlying features. $L=100,000$, $M=80,000$, $|{\cal S}|=30,000$, $\sigma=1$, $I=3$.}
   	\resizebox{.65\linewidth}{!}{\huge
 \begin{tabular}{c|c|c|c|c|c}
     \hline
     \hline
     \multirow{2}[4]{*}{Algorithm} & \multirow{2}[4]{*}{measure} & \multicolumn{4}{c}{$\rho$ (density of views)} \\
 \cline{3-6}          &       & $10^{-5}$ & $5\times 10^{-4}$ & $10^{-4}$ & $10^{-3}$ \\
     \hline
     \hline
     \multirow{3}[6]{*}{\texttt{MVLSA} ($P=100$)} & metric1 & 16.843 & 13.877 & 17.159 & 16.912 \\
 \cline{2-6}          & metric2 & \textbf{0.003} & \textbf{0.010} & 0.009 & 0.003 \\
 \cline{2-6}          & time (min) & \textbf{0.913} & \textbf{1.019} & \textbf{1.252} & \textbf{3.983} \\
     \hline
     \hline
     \multirow{3}[6]{*}{Proposed ($\mu=.05$)} & metric1 & \textbf{0.478} & \textbf{0.610} & \textbf{0.565} & \textbf{0.775} \\
 \cline{2-6}          & metric2 & 0.018 & 0.134 & 0.034 & 0.003 \\
 \cline{2-6}          & time  (min) & 3.798	&5.425&	5.765&	24.182 \\
     \hline
     \hline
     \multirow{3}[6]{*}{Proposed ($\mu=.1$)} & metric1 & 0.942 & 1.054 & 0.941 & 1.265 \\
 \cline{2-6}          & metric2 & 0.006 & 0.054 & \textbf{0.004} & \textbf{0.000} \\
 \cline{2-6}          & time  (min) & 2.182&	3.791&	4.510&	16.378 \\
     \hline
     \hline
     \multirow{3}[6]{*}{Proposed ($\mu=.5$)} & metric1 & 1.592 & 1.497 & 1.306 & 1.538 \\
 \cline{2-6}          & metric2 & \textbf{0.003} & 0.021 & \textbf{0.000} & \textbf{0.000} \\
 \cline{2-6}          & time  (min) & 1.735	&2.714&	3.723	&13.447\\
     \hline
     \hline
     \end{tabular}%
   	}%
  \label{tab:rhochange}%
\end{table}%

\begin{table}[htbp]
  \centering
    \caption{Evaluation of the algorithm versus the number of views in the presence of outlying features. $L=100,000$, $M=80,000$, $|{\cal S}|=30,000$, $\sigma=1$, $\rho=5\times 10^{-5}$.}
   \resizebox{.65\linewidth}{!}{\huge
       \begin{tabular}{c|c|c|c|c|c|c|c}
       \hline
       \hline
       \multirow{2}[4]{*}{Algorithm} & \multirow{2}[4]{*}{measure} & \multicolumn{6}{c}{$I$ (no. of views)} \\
   \cline{3-8}          &       & 3  & 4  & 5 & 6  & 7  & 8 \\
       \hline
       \hline
       \multirow{3}[6]{*}{\texttt{MVLSA} ($P=100$)} & metric1 & 15.813 & 15.715 & 14.667 & 16.904 & 17.838 & 17.691 \\
   \cline{2-8}          & metric2 & \textbf{0.008} & 0.009 & 0.009 & 0.009 & 0.007 & 0.009 \\
   \cline{2-8}          & time (min) & \textbf{1.087} & \textbf{0.975} & \textbf{0.960} & \textbf{0.958} & \textbf{0.989} & \textbf{1.026} \\
       \hline
       \hline
       \multirow{3}[6]{*}{proposed ($\mu=.05$)} & metric1 & \textbf{0.731} & \textbf{0.590} & \textbf{0.670} & \textbf{0.611} & \textbf{0.517} & \textbf{0.628} \\
   \cline{2-8}          & metric2 & 0.172 & 0.078 & 0.100 & 0.101 & 0.065 & 0.098 \\
   \cline{2-8}          & time (min) & 5.870&	6.064&	5.762&	5.070&	5.895&	5.776  \\
       \hline
       \hline
       \multirow{3}[6]{*}{proposed ($\mu=.1$)} & metric1 & 1.070 & 1.057 & 1.110 & 1.026 & 1.042 & 1.112 \\
   \cline{2-8}          & metric2 & 0.055 & 0.019 & 0.018 & 0.024 & 0.023 & 0.023 \\
   \cline{2-8}          & time (min) & 3.240&	2.974&	3.313&	3.210&	3.083&	3.529\\
       \hline
       \hline
       \multirow{3}[6]{*}{proposed ($\mu=.5$)} & metric1 & 1.461 & 1.482 & 1.578 & 1.443 & 1.472 & 1.561 \\
   \cline{2-8}          & metric2 & 0.018 & \textbf{0.002} & \textbf{0.003} & \textbf{0.001} & \textbf{0.006} & \textbf{0.007} \\
   \cline{2-8}          & time (min) & 2.700&	2.441&	2.528&	2.569&	2.431&	2.567 \\
       \hline
       \hline
       \end{tabular}%
     \label{tab:addlabel}%
   }%
  \label{tab:Ichange}%
\end{table}%

\subsection{Real Large-Scale Word Embedding Tasks}
We test the algorithms on a large-scale multilingual dataset.
The views are extracted from a large word co-occurrence matrix, which is available at \url{https://sites.google.com/a/umn.edu/huang663/research}.
The original data contains words of three languages, namely, English, Spanish, and French,
and all the words are defined by the co-occurences pointwise mutual information (PMI) with other words.
We use the English words to form our first view, ${\bm X}_1$, which contains $L=183,034$ words and each word is defined by $M_i=100,000$ features (co-occurrences). Note that ${\bm X}_1$ is sparse -- only $1.21\%$ of its entries are non-zeros.
Using a dictionary, we pick out the translations of the English words contained in ${\bm X}_1$ in Spanish and French to form ${\bm X}_2$ and ${\bm X}_3$, respectively.
Note that many English words do not have a corresponding word in Spanish (or French).
In such cases, we simply let ${\bm X}_i(\ell,:)={\bm 0}$ for $i=2$ (or $i=3$),
resulting in sparser ${\bm X}_2$ and ${\bm X}_3$.
Our objective is to use ``side information'' provided by Spanish and French to find a ${\bm G}$ whose rows are low-dimensional embeddings of the English words (cf. the motivating example in Fig.~\ref{fig:motivation}).

To evaluate the output, we use the evaluation tool provided at {wordvectors.org} \cite{faruqui-2014:SystemDemo}, which runs several word embedding tasks to evaluate a set of given embeddings. Simply speaking, the tasks compare the algorithm-learned
embeddings with the judgment of humans and yield high scores if the embeddings are consistent with the humans. The scores are between zero and one,
and a score equal to one means a perfect alignment between the learned result and human judgment.
We use the result of \texttt{MVLSA} with $P=640$ as benchmark.
The result of applying SVD to ${\bm X}_1$ without considering different languages is also presented.
We apply the proposed algorithm warm started by \texttt{MVLSA} and set $T=1$.
We run {three} versions of our algorithm. The first one uses $h_i(\cdot)=\frac{\mu_i}{2}\|\cdot\|_F^2$ with $\mu_i=1$
for $i=1,2,3$.
The second one uses $h_i(\cdot)=\mu_i\|\cdot\|_{2,1}$ for $i=2,3$ where $\mu_i=0.05$, and we have no regularization on the first view.
The third one is $h_i(\cdot)=\mu_i\|\cdot\|_{1,1}$ for $i=2,3$ where $\mu_i=0.05$.
The reason for adding $\ell_2/\ell_1$ mixed-norm (and $\ell_1$ norm) regularization to the French and Spanish views
is twofold:
First, the $\ell_2/\ell_1$ norm ($\ell_1$ norm) promotes row sparsity (sparsity) of $\Q_i$ and thus performs feature selection on $\X_2$ and $\X_3$ -- this physically means that we aim at selecting the most useful features from the other languages to help enhance English word embeddings.
Second, $\X_2$ and $\X_3$ are effectively ``fat matrices'' and thus a column-selective regularizer can help improve the conditioning.
Interestingly, we find that not adding feature-selective regularizations to the English view produces better results for the dataset considered. Our understanding is that $\X_1$ is a complete view without missing elements, and thus giving $\Q_1$ more ``degrees of freedom'' helps improve performance.

Tables~\ref{tab:K50} and \ref{tab:K100} show the word embedding results using $K=50$ and $K=100$, respectively.
One can see that using the information from multiple views does help in improving the word embeddings:
For $K=50$ and $K=100$, the multiview approaches perform better relative to SVD in 11 and 12 tasks out of 12 tasks.
In addition, the proposed algorithm with the regularizer $h_i(\cdot)=\nicefrac{\mu_i}{2}\|\cdot\|_F^2$ (denoted by $\ell_2$) gives similar or slightly better performance on average in both experiments compared to \texttt{MVLSA}.
The proposed algorithm with the feature-selective regularizers (denoted by $\ell_2/\ell_1$ and $\ell_1$, resp.) gives the best evaluation results on both experiments -- this suggests that for large-scale multilingual word embedding, feature selection is very meaningful.
In particular, we observe that using $h_i(\Q_i)=\|\Q_i\|_{1,1}$ gives the best performance on many tasks. This further suggests that, in this case, different components of the reduced-dimension representations (i.e., columns of $\X_i\Q_i$) may be better learned by using different features of the views.


\begin{table*}[htbp]
  \centering
  \caption{Evaluation on 12 word embedding tasks; $K=50$.}
  	\resizebox{.7\linewidth}{!}{
        \begin{tabular}{c|c|c|c|c|c}
        \hline
        \hline
        \multirow{2}[4]{*}{Task} & \multicolumn{5}{c}{Algorithm ($K=50$)}\\
    \cline{2-6}          & SVD   & \texttt{MVLSA} & \texttt{AltMaxVar} ($\ell_2$) & \texttt{AltMaxVar}  ($\ell_2/\ell_1$) & \texttt{AltMaxVar} ($\ell_1$)\\
        \hline
        \hline
        EN-WS-353-SIM & 0.63  & {\bf 0.69} & 0.67  & 0.68  & {\bf 0.69}\\
        \hline
        EN-MC-30 & 0.56  & 0.63  & 0.63  & {0.64}&  {\bf 0.66}\\
        \hline
        EN-MTurk-771 & 0.54  & 0.58  & 0.59  & {\bf 0.60}& 0.59\\
        \hline
        EN-MEN-TR-3k & 0.67  & 0.66  & 0.67  & {0.68}&{\bf 0.69}\\
        \hline
        EN-RG-65 & 0.51  & 0.53  & 0.55  & {\bf 0.58}&{\bf 0.58}\\
        \hline
        EN-MTurk-287 & {\bf 0.65} & 0.64  & {\bf 0.65} & 0.64&0.63\\
        \hline
        EN-WS-353-REL & 0.50  & 0.51  & 0.53  & {0.55} & {\bf 0.57}\\
        \hline
        EN-VERB-143 & 0.21  & {\bf 0.22} & 0.21  & 0.21 & 0.20\\
        \hline
        EN-YP-130 & 0.36  & 0.39  & 0.38  & {\bf 0.41} & {\bf 0.41}\\
        \hline
        EN-SIMLEX-999 & 0.31  & {\bf 0.42} & 0.41  & 0.39 & 0.36\\
        \hline
        EN-RW-STANFORD & 0.39  & {\bf 0.43} & {\bf 0.43} & {\bf 0.43}& {\bf 0.43}\\
        \hline
        EN-WS-353-ALL & 0.56  & 0.59  & 0.59  & {0.60} & {\bf 0.62}\\
        \hline
        \hline
        Average & 0.49  & 0.52  & 0.53  & {\bf 0.54} &{\bf 0.54}\\
        \hline
        Median & 0.53  & 0.56  & 0.57  & {\bf 0.59} & {\bf 0.59}\\
        \hline
        \hline
        \end{tabular}%

    }%
  \label{tab:K50}%
\end{table*}%

\begin{table*}[htbp]
  \centering
  \caption{Evaluation on 12 word embedding tasks; $K=100$.}
  	\resizebox{.7\linewidth}{!}{
    \begin{tabular}{c|c|c|c|c|c}
        \hline
        \hline
        \multirow{2}[4]{*}{Task} & \multicolumn{5}{c}{Algorithm ($K=100$)}\\
        \cline{2-6}          & SVD   & \texttt{MVLSA} & \texttt{AltMaxVar} ($\ell_2$) & \texttt{AltMaxVar}  ($\ell_2/\ell_1$) & \texttt{AltMaxVar} ($\ell_1$)\\
        \hline
        \hline
    EN-WS-353-SIM & 0.68  & {\bf 0.72} & 0.71  & {\bf 0.72} & {\bf 0.72}\\
    \hline
    EN-MC-30 & 0.73  & 0.68  & 0.72  & {0.74} & {\bf 0.82}\\
    \hline
    EN-MTurk-771 & 0.59  & 0.60  & 0.60  & {0.61} &{\bf 0.62}\\
    \hline
    EN-MEN-TR-3k & {0.72} & 0.70  & 0.70  & 0.71 & {\bf 0.73}\\
    \hline
    EN-RG-65 & {0.68} & 0.63  & 0.64  & {0.68} &  {\bf 0.70}\\
    \hline
    EN-MTurk-287 & 0.61  & {\bf 0.66} & 0.65  & 0.64 & 0.64\\
    \hline
    EN-WS-353-REL & {0.57} & 0.54  & 0.55  & 0.56 & {\bf 0.59}\\
    \hline
    EN-VERB-143 & 0.19  & 0.28  & 0.27  & {\bf 0.29} & {0.28}\\
    \hline
    EN-YP-130 & 0.42  & 0.41  & 0.41  & {0.45} & {\bf 0.49}\\
    \hline
    EN-SIMLEX-999 & 0.34  & {\bf 0.42} & 0.41  & 0.41 & 0.39\\
    \hline
    EN-RW-STANFORD & 0.44  & {0.46} & 0.45  & {0.46} & {\bf 0.48}\\
    \hline
    EN-WS-353-ALL & {0.62} & {0.62} & {0.62} & {0.62} & {\bf 0.65}\\
    \hline
    \hline
    Average & 0.55  & 0.56  & 0.56  & {0.58} & {\bf 0.59}\\
    \hline
    Median & 0.60  & 0.61  & 0.61  & {0.62} & {\bf 0.63}\\
    \hline
    \hline
    \end{tabular}%
    }
  \label{tab:K100}%
\end{table*}%

\section{Conclusion and Future Work}
In this work, we revisited the MAX-VAR GCCA problem with an eye towards scenarios involving large-scale and sparse data.
The proposed approach is memory-efficient and has light per-iteration computational complexity if the views are sparse, and is thus suitable for dealing with big data.
The algorithm is also flexible for incorporating different structure-promoting regularizers on the canonical components such as feature-selective regularizations. A thorough convergence analysis was presented, showing that the proposed algorithmic framework guarantees a KKT point to be obtained at a sublinear convergence rate in general cases under a variety of structure-promoting regularizers.
We also showed that the algorithm approaches a global optimal solution at a linear convergence rate if the original MAX-VAR problem without regularization is considered. Simulations and real experiments with large-scale multi-lingual data showed that the performance of the proposed algorithm is promising in dealing with real-world large and sparse multiview data. 

In the future, it is interesting to consider nonlinear operator-based multiview analysis, e.g., kernel (G)CCA \cite{hardoon2004canonical} or deep neural network-based (G)CCA \cite{andrew2013deep}, under large-scale settings. Nonlinear dimensionality reduction is very well-motivated in practice since it is able to handle more complex models and usually performs well with real-life data. On the other hand, the associated optimization problems are much harder, especially when the data dimension is large -- which also promises a fertile research ground ahead. Another interesting direction is to consider constraints on $\G$ (or $\X_i\Q_i$) -- in some applications, structured (e.g., sparse and nonnegative) low-dimensional representations of data are desired.

\ifplainver
    \section*{Appendix}
    \renewcommand{\thesubsection}{\Alph{subsection}}
\else
\appendices
\fi

\section{Proof of Proposition~\ref{lem:monotonicity}}
To simplify the notation, let us define ${\bm Q}=[{\bm Q}_1^T,\ldots,{\bm Q}_I^T]^T$ as a collection
of $\Q_i$'s. 
We also rewrite the objective function in \eqref{eq:CGCCA_reg} as 
\begin{align*}
F(\Q,\G)&=f(\Q,\G)+g(\Q)\\
&=\sum_{i=1}^I f_i(\Q_i,\G)+\sum_{i=1}^I g_i(\Q_i),
\end{align*}
where $f_i(\Q_i,\G)$ and $g_i(\Q_i)$ are the smooth and non-smooth parts in \eqref{eq:Qi} as before,
and $f(\Q,\G)=\sum_{i=1}^I f_i(\Q_i,\G)$ and $g(\Q) = \sum_{i=1}^I g_i(\Q_i)$, respectively.
Additionally, let
\begin{align*}
\nabla_{\Q}~f(\Q,\G)& =[(\nabla_{\Q_1}f(\Q,\G))^T,\ldots,(\nabla_{\Q_I}f(\Q,\G))^T]^T,\\
\partial_{\Q} g(\Q) &=[( \partial_{\Q_1} g_1(\Q_1))^T,\ldots,(\partial_{\Q_I} g_I(\Q_I))^T]^T.
\end{align*}
As the algorithm is essentially a two-block alternating optimization (since ${\bm Q}_i$ for all $i$ are updated simultaneously), the above notation suffices to describe the updates.
Let
\begin{align*}
u_Q\left(\Q;\hat{\bm G},\hat{\bm Q}\right) = &f(\hat{\bm G},\hat{\bm Q}) +  \left<\nabla_{{\bm Q}} f(\hat{\bm Q},\hat{\bm G}),{\bm Q}-\hat{\bm Q}\right>\\
&+ \sum_{i=1}^I\frac{1}{2\alpha_i}\|{\bm Q}_i-\hat{\bm Q}\|_F^2+ \sum_{i=1}^Ig_i(\Q_i);
\end{align*}
i.e., $u_Q\left({\bm Q};\hat{\bm G},\hat{\bm Q}\right)$ is an approximation of $F({\bm G},{\bm Q})$
locally at the point $(\hat{\bm G},\hat{\bm Q})$.
We further define
$\tilde{u}_Q\left(\Q;\hat{\bm G},\hat{\bm Q}\right) = u_Q\left({\bm Q};\hat{\bm G},\hat{\bm Q}\right) -\sum_{i=1}^Ig_i(\Q_i)$;
i.e., $\tilde{u}_Q\left(\Q;\hat{\bm G},\hat{\bm Q}\right)$ is an approximation of the continuously differentiable part $f({\bm G},{\bm Q})$
locally at the point $(\hat{\bm G},\hat{\bm Q})$.
One can see that,
\begin{align}\label{eq:gradequal}
	&\nabla_{{\bm Q}} f\left(\hat{\bm Q},\hat{\bm G}\right)=\nabla_{{\bm Q}} \tilde{u}\left(\hat{\bm Q};\hat{\bm G},\hat{\bm Q}\right),
\end{align}
Since $\nabla_{{\bm Q}_i} f_i({\bm Q}_i,\G)$ is $L_i$-Lipschitz continuous w.r.t. ${\bm Q}_i$ and $\alpha_i\leq 1/L_i$ for all $i$, we have the following holds:
\begin{equation}\label{eq:gleqf}
	u_Q\left({\bm Q};\hat{\bm G},\hat{\bm Q}\right)\geq F\left({\bm Q},\hat{\bm G}\right),~\forall~{\bm Q},
\end{equation}
where the equality holds if and only if ${\bm Q}_i = \hat{\bm Q}_i$ for all $i$, i.e.,
\begin{equation}\label{eq:geqf}
u_Q\left(\hat{\bm Q};\hat{\bm G},\hat{\bm Q}\right) = F\left(\hat{\bm Q},\hat{\bm G}\right).
\end{equation}
Similarly, we define
\begin{align*}
u_G\left(\G;\hat{\bm G},\hat{\bm Q}\right) = &\sum_{i=1}^I\frac{1}{2}\left\|{\bm X}_i{\bm Q}_i^{(r+1)}-{\bm G}\right\|_F^2\\
&+\omega\cdot\left\|{\bm G}-{\bm G}^{(r)}\right\|_F^2 + \sum_{i=1}^Ig_i(\Q_i),
\end{align*}
where we recall that $\omega = \nicefrac{(1-\gamma)I}{2\gamma}$ and the last term is a constant if $\Q$ is fixed.

The update rule of $\G$ in Algorithm~\ref{algo:AltCCA} can be re-expressed as $
\G  \in  \arg\min_{\G^T\G={\bm I}}~u_G\left(\G;\hat{\bm G},\hat{\bm Q}\right)$.
It is easily seen that
\begin{subequations}\label{eq:ugleqf}
\begin{align}
&	u_G\left({\bm G};\hat{\bm G},\hat{\bm Q}\right)\geq F\left(\hat{\bm Q},{\bm G}\right),\\
&	u_G\left(\hat{\bm G};\hat{\bm G},\hat{\bm Q}\right)= F\left(\hat{\bm Q},\hat{\bm G}\right).
\end{align}
\end{subequations}
Hence, Algorithm~\ref{algo:AltCCA} boils down to
\begin{subequations}
\begin{align}
{\bm Q}_i^{(r,t+1)}&= \arg\min_{{\bm Q}_i}~u_{Q}\left( {\bm Q};{\bm G}^{(r)},{\bm Q}^{(r,t)} \right),\quad \forall t\label{eq:q_update_u}\\
\G^{(r+1)} &\in \arg\min_{\G^T\G={\bm I}}~u_G\left(\G;{\bm G}^{(r)},{\bm Q}^{(r+1)}\right). \label{eq:g_update_u}
\end{align}
\end{subequations}
When $\gamma=1$, \eqref{eq:g_update_u} amounts to SVD of $\sum_{i=1}^I\X_i\Q_i/I$ and the ${\bm G}$-subproblem $\min_{\G^T\G={\bm I}}F({\bm Q}^{(r+1)},\G)$ is optimally solved;
otherwise, both \eqref{eq:q_update_u} and \eqref{eq:g_update_u} are local upper bound minimizations.

Note that the following holds:
\begin{subequations}\label{eq:mono}
\begin{align}
         F\left({\bm Q}^{(r)},{\bm G}^{(r)}\right) & = u_Q(\Q^{(r)};{\bm G}^{(r)},{\bm Q}^{(r)}) \label{eq:mono1}\\
                                                   & \geq u_Q(\Q^{(r+1)};{\bm G}^{(r)},{\bm Q}^{(r,T-1)}) \label{eq:mono3}\\
                                                   &\geq F\left({\bm Q}^{(r+1)},{\bm G}^{(r)}\right) \label{eq:mono4}\\
                                                   &= u_G\left({\bm G}^{(r)};{\bm G}^{(r)},{\bm Q}^{(r+1)}\right) \label{eq:mono5}\\
												   &\geq u_G\left({\bm G}^{(r+1)};{\bm G}^{(r)},{\bm Q}^{(r+1)}\right) \label{eq:mono6}\\
												   &\geq F\left({\bm Q}^{(r+1)},{\bm G}^{(r+1)}\right), \label{eq:mono7}
\end{align}
\end{subequations}
where \eqref{eq:mono1} holds because of \eqref{eq:geqf}, \eqref{eq:mono3} holds
since PG is a descending method when $\alpha_i\leq 1/L_i$ \cite{beck2009fast},
\eqref{eq:mono4} holds by the property in \eqref{eq:geqf},
\eqref{eq:mono5} holds due to \eqref{eq:ugleqf},
\eqref{eq:mono6} is due to the fact that \eqref{eq:g_update_u} is optimally solved,
and \eqref{eq:mono7} holds also because of the first equation in \eqref{eq:ugleqf}.

Next, we show that every limit point is a KKT point.
Assume that there exists a convergent subsequence of $\{{\bm G}^{(r)},{\bm Q}^{(r)}\}_{r=0,1,\ldots}$,
whose limit point is $({\bm G}^\ast,{\bm Q}^\ast)$ and the subsequence is indexed by $\{r_j\}_{j=1,\ldots,\infty}$.
We have the following chain of inequalities:
\begin{subequations}\label{eq:u_Q}
\begin{align}
         u_Q\left({\bm Q};{\bm G}^{(r_j)},{\bm Q}^{(r_j)}\right) &\geq u_Q\left({\bm Q}^{(r_j,1)};{\bm G}^{(r_j)},{\bm Q}^{(r_j)}\right) \label{eq:cmin1}\\
				                               &\geq u_Q\left({\bm Q}^{(r_j,T)};{\bm G}^{(r_j)},{\bm Q}^{(r_j,T-1)}\right) \label{eq:cmin23}\\
				                              &\geq F({\bm G}^{(r_j)},{\bm Q}^{(r_j+1)})\label{eq:cmin24}\\					
                                               &\geq F\left({\bm Q}^{(r_j+1)},{\bm G}^{(r_j+1)}\right)  \label{eq:cmin3}\\
											  &\geq F\left({\bm Q}^{(r_{j+1})},{\bm G}^{(r_{j+1})}\right)  \label{eq:cmin4}\\
											   & = u_Q\left({\bm Q}^{(r_{j+1})};{\bm G}^{(r_{j+1})},{\bm Q}^{(r_{j+1})}\right), \label{eq:cmin5}
\end{align}
\end{subequations}
where \eqref{eq:cmin1} holds because of the update rule in \eqref{eq:q_update_u},
\eqref{eq:cmin23} holds, again, by the descending property of PG,
\eqref{eq:cmin3} follows \eqref{eq:mono7},
and \eqref{eq:cmin5} is again because of the way that we construct $u_Q({\bm Q};{\bm G}^{(r_{j+1)}},{\bm Q}^{(r_{j+1})})$.
Taking $j\rightarrow \infty$, and by continuity of $u_Q(\cdot)$, we have
\begin{equation}
	    u_Q({\bm Q};{\bm G}^{\ast},{\bm Q}^\ast) \geq  u_Q({\bm Q}^\ast;{\bm G}^{\ast},{\bm Q}^\ast),
\end{equation}
i.e., ${\bm Q}^{\ast}$ is a minimum of $u_Q({\bm Q};{\bm G}^{\ast},{\bm Q}^\ast)$.
Consequently, ${\bm Q}^{\ast}$ satisfies the conditional KKT conditions, i.e.,
$	 {\bm 0} \in  \nabla_{{\bm Q}} \tilde{u}_Q({\bm Q}^\ast;{\bm G}^{\ast},{\bm Q}^\ast) + \partial_{\Q}g(\Q^\ast),$
which, by \eqref{eq:gradequal}, also means
\begin{equation}\label{eq:QKKT}
	     {\bm 0} \in \nabla_{{\bm Q}_i} f_i({\bm Q}_i^\ast,{\bm G}^{\ast}) + \partial_{\Q_i}g_i(\Q_i^\ast),~\forall i.
\end{equation}

We now show that $\Q^{(r_j,t)}$ for $t=1,\ldots,T$ also converges to $\Q^\ast$.
Indeed, we have
\begin{align*}
u_Q(\Q^{(r_{j+1})};\G^{(r_{j+1})},\Q^{(r_{j+1})})&\leq u_Q(\Q^{(r_j,1)};\G^{(r_j)},\Q^{(r_j)})\\
&\leq u_Q(\Q^{(r_j)};\G^{(r_j)},\Q^{(r_j)}) ,
\end{align*}
where the first inequality was derived from \eqref{eq:u_Q}.
Taking $j\rightarrow \infty$, we see that
$ u_Q(\Q^{\ast};\G^{\ast},\Q^{\ast})\leq u_Q(\Q^{(r_j,1)};\G^{\ast},\Q^{\ast})\leq u_Q(\Q^{\ast};\G^{\ast},\Q^{\ast}) , $
which implies that
$u_Q(\Q^{(r_j,1)};\G^{\ast},\Q^{\ast})= u_Q(\Q^{\ast};\G^{\ast},\Q^{\ast})\leq u_Q(\Q;\G^{\ast},\Q^{\ast}).$
On the other hand, the problem in \eqref{eq:q_update_u} has a unique minimizer when $g_i(\cdot)$ is a convex closed function \cite{parikh2013proximal},
which means that $\Q^{(r_j,1)}\rightarrow \Q^{\ast}$. By the same argument, we can show that
$\Q^{(r_j,t)}$ for $t=1,\ldots,T$ also converges to $\Q^\ast$.
Consequently, we have
$\Q^{(r_j,T)} = \Q^{(r_j+1)} \rightarrow \Q^\ast$.
We repeat the proof in \eqref{eq:u_Q} to ${\bm G}$:
\begin{subequations}\label{eq:u_G}
\begin{align*}
         u_G\left({\bm G};{\bm G}^{(r_j)},{\bm Q}^{(r_j+1)}\right) &\geq u_G\left({\bm G}^{(r_j+1)};{\bm G}^{(r_j)},{\bm Q}^{(r_j+1)}\right) \\
				                              &\geq F({\bm Q}^{(r_j+1)},{\bm G}^{(r_j+1)})\\					
                                               &\geq F\left({\bm Q}^{(r_j+1)},{\bm G}^{(r_j+1)}\right)  \\
											   & = u_G\left({\bm G}^{(r_{j+1})};{\bm G}^{(r_{j+1})},{\bm Q}^{(r_{j+1})}\right),
\end{align*}
\end{subequations}
Taking $j\rightarrow \infty$ and by $\Q^{(r_j+1)} \rightarrow \Q^\ast$, we have
\[ u_G\left({\bm G};{\bm G}^{\ast},{\bm Q}^{\ast}\right) \geq u_G\left({\bm G}^{\ast};{\bm G}^{\ast},{\bm Q}^{\ast}\right),\quad \forall \G^T\G={\bm I}. \]
The above means that $\G^\ast$ satisfies the partial conditional KKT conditions w.r.t. $\G$.
Combining with \eqref{eq:QKKT}, we see that $({\bm G}^\ast,{\bm Q}^\ast)$ is a KKT point of the original problem.

\smallskip

Now, we show the b) part.
First, we show that ${\bm Q}_i$ remains in a bounded set (the variable ${\bm G}$ is always bounded since we keep it feasible in each iteration).
Since the objective value is non-increasing (cf. Proposition~\ref{lem:monotonicity}), if we denote the initial objective value as $V$, then $F({\bm G}^{(r)},{\bm Q}^{(r)}) \leq V$ holds in all subsequent iterations.
Note that when ${\bm X}_i^{(0)}$ and ${\bm Q}_i^{(0)}$ are bounded, $V$ is also finite.
In particular, we have $\left\| {\bm X}_i {\bm Q}_i - {\bm G} \right\|_F^2 + 2\sum_{i=1}^Ig_i(\Q_i) \leq 2V$
holds, which implies
$ \| {\bm X}_i {\bm Q}_i \|_F \leq \| {\bm G} \|_F + \sqrt{2V}$
by the triangle inequality.
The right-hand side is finite since both terms are bounded. Denote $ ( \| {\bm G} \|_F + \sqrt{2V} )$ by $V'$. Then, we have
$\| {\bm Q}_i \|_F = \| ({\bm X}_i^T{\bm X}_i)^{-1} {\bm X}_i^T{\bm X}_i {\bm Q}_i \|_F
\leq \| ({\bm X}_i^T{\bm X}_i)^{-1} {\bm X}_i^T \|_F \cdot \|{\bm X}_i {\bm Q}_i \|_F
\leq V' \cdot \| ({\bm X}_i^T{\bm X}_i)^{-1} {\bm X}_i^T \|_F.$
Now, by the assumption that ${\rm rank}({\bm X}_i)=M_i$, the term $\| ({\bm X}_i^T{\bm X}_i)^{-1} {\bm X}_i^T \|_F$ is bounded.  This shows that $\| {\bm Q}_i \|_F$ is bounded.
Hence, starting from a bounded ${\bm Q}_i^{(0)}$, the solution sequence $\{{\bm Q}{(r)},{\bm G}^{(r)}\}$ remains in a bounded set. Since the constraints of ${\bm Q}_i$, i.e., $\mathbb{R}^{M_i\times K}$ and ${\bm G}$ are also closed sets,  $\{{\bm Q}^{(r)},{\bm G}^{(r)}\}$ remains in a compact set.

Now, let us denote ${\cal K}$ as the set containing all the KKT points.
Suppose the whole sequence does not converge to ${\cal K}$.
Then, there exists a convergent subsequence indexed by $\{r_j\}$ such that
$\lim_{j\rightarrow \infty} d^{(r)}({\cal K})\geq \gamma$
for some positive $\gamma$, where
$d^{(r)}({\cal K}) = \min_{{\bm Y}\in{\cal K}}~\|({\bm G}^{(r)},{\bm Q}^{(r)}) - {\bm Y}\|.$
Since the subsequence indexed by $\{r_j\}$ lies in a closed and bounded set as we have shown,
this subsequence has a limit point.
However, as we have shown in Theorem~\ref{lem:monotonicity}, every limit point of the solution sequence is a KKT point.
This is a contradiction.
Therefore, the whole sequence converges to a KKT point.

\section{Proof of Lemma~\ref{lem:z}}\label{app:lemma_z}
First, we have the update rule
$\Q_i^{(r,t+1)} = \Q_i^{(r,t)}-\alpha_i\tilde{\nabla}_{\Q_i}F(\Q_i^{(r,t)},\G^{(r)}),$
which leads to the following:
\begin{equation}\label{eq:q_norm}
\frac{1}{\alpha_i}(\Q_i^{(r,t+1)} - \Q_i^{(r,t)} )= -\tilde{\nabla}_{\Q_i}F(\Q_i^{(r,t)},\G^{(r)}).
\end{equation}
Meanwhile, the updating rule can also be expressed as
\begin{align}
\Q_i^{(r,t+1)} = &\arg\min_{\Q_i}~ \left<\nabla_{\Q_i}f(\Q_i^{(r,t)},\G^{(r)}),\Q_i -\Q_i^{(r,t)}\right> \nonumber\\
     & +g_i(\Q_i)+\frac{1}{2\alpha_i}\|\Q_i-\Q_i^{(r,t)}\|_F^2.  \label{eq:q_arg}
\end{align}
Therefore,
there exists a $\partial_{\Q_i}g_i(\Q^{(r,t+1)})$ and a $\Q^{(r,t+1)}$ satisfy the following optimality conditions:
\begin{align*}
&{\bm 0} = \nabla_{\Q_i}f_i(\Q_i^{(r,t)},\G^{(r)}) + \partial_{\Q_i}g_i(\Q_i^{(r,t+1)})+\frac{1}{\alpha_i}(\Q_i^{(r,t+1)}-\Q_i^{(r,t)}).
\end{align*}
Consequently, we see that
\begin{align*}
&\sum_{i=1}^I\sum_{t=0}^T\left\|\tilde{\nabla}_{\Q_i}F(\Q_i^{(r,t)},\G^{(r)})\right\|_F^2\rightarrow 0 \\
&\Rightarrow \Q_i^{(r,t)}-\Q_i^{(r,t+1)} \rightarrow {\bm 0},~\forall~t=0,\ldots,T-1\\
&\Rightarrow \Q_i^{(r)}-\Q_i^{(r+1)} \rightarrow 0,~\forall i\\
&\Rightarrow \nabla_{\Q}~f\left(\Q^{(r)},\G^{(r)}\right) + \partial_{\Q} g\left(\Q^{(r)}\right)  \rightarrow 0 \nonumber
\end{align*}
which holds since $T$ is finite.
The above means that
${\bm 0}\in \nabla_{\Q}f(\Q^{(r)},\G^{(r)}) + \partial_{\Q}g(\Q^{(r)})$
is satisfied when $Z^{(r+1)}\rightarrow 0$.


Recall that $\G^{(r+1)}$ satisfies the optimality condition of Problem~\eqref{eq:G-sub-prox}.
Therefore, there exists a ${\bm \Lambda}^{(r+1)}$ such that
the following optimality condition holds
\begin{align}
&\G^{(r)} -\sum_{i=1}^I\X_i\Q_i^{(r+1)} /I+\frac{1}{{\gamma}}\left(\G^{(r+1)} -\G^{(r)}\right) \nonumber\\
& \quad+ \G^{(r+1)}{\bm\Lambda}^{(r+1)}={\bm 0} \label{eq:g_norm}
\end{align}
Combining \eqref{eq:g_norm} and \eqref{eq:q_norm}, we have
\begin{equation*}
Z^{(r+1)} = \frac{1}{{\gamma}^2}  \left\|\G^{(r+1)} -\G^{(r)}\right\|_F^2 + \sum_{i=1}^I\frac{1}{\alpha_i^2}\left\|\Q_i^{(r+1)}-\Q_i^{(r)}\right\|_F^2.
\end{equation*}
We see that $Z^{(r+1)}\rightarrow 0$ implies that a KKT point is reached and this completes the proof of Lemma~\ref{lem:z}.

\section{Proof of Theorem~\ref{thm:complexity}}\label{app:complexity}

We show that every iterate of $\Q$ and $\G$ gives sufficient decreases of the overall objective function.
Since $\nabla_{\Q_i}f_i(\Q_i,\G)$ is $L_i$-Lipschitz continuous for all $i$, we have the following:
\begin{align} \label{eq:24}
&F(\Q^{(r,t+1)},\G^{(r)})\leq  u_Q\left(\Q^{(r,t+1)};{\bm G}^{(r)},{\bm Q}^{(r,t)}\right).\\
&=f(\Q^{(r,t)},\G^{(r)}) +  \left<\nabla_{\Q}f(\Q^{(r,t)},\G^{(r)}),\Q^{(r,t+1)} -\Q^{(r)}\right> \nonumber\\
&  + \sum_{i=1}^Ig_i\left(\Q_i^{(r,t+1)}\right) + \sum_{i=1}^I\frac{L_i}{2}\left\|\Q_i^{(r,t+1)}-\Q_i^{(r,t)}\right\|_F^2. \nonumber
\end{align}
Since $\Q^{(r,t+1)}$ is a minimizer of Problem~\eqref{eq:q_arg}, we also have
\begin{align}\label{eq:25}
 &\left<\nabla_{\Q}f(\Q^{(r,t)},\G^{(r)}),\Q^{(r,t+1)} -\Q^{(r,t)}\right>+\sum_{i=1}^Ig_i(\Q_i^{(r,t+1)}) \nonumber\\
  &+ \sum_{i=1}^I\frac{1}{2\alpha_i}\left\|\Q_i^{(r,t+1)}-\Q_i^{(r,t)}\right\|_F^2 \leq  \sum_{i=1}^Ig_i(\Q_i^{(r,t)}),
\end{align}
which is obtained by letting $\Q_i=\Q_i^{(r,t)}$.
Combining \eqref{eq:24} and \eqref{eq:25}, we have
\begin{equation}\label{eq:suff_q0}
\begin{aligned}
&F(\Q^{(r,t+1)},\G^{(r)}) - F(\Q^{(r,t)},\G^{(r)}) \\
& \leq -\sum_{i=1}^I\left( \frac{1}{2\alpha_i} - \frac{L_i}{2} \right)\left\|\Q_i^{(r,t+1)}-\Q_i^{(r,t)}\right\|_F^2.
\end{aligned}
\end{equation}
Summing up the above over $t=0,\ldots,T-1$, we have
\begin{equation}\label{eq:suff_q}
\begin{aligned}
& F(\Q^{(r)},\G^{(r)}) - F(\Q^{(r+1)},\G^{(r)})\\
&\geq \sum_{t=0}^{T-1}\sum_{i=1}^I\left( \frac{1}{2\alpha_i} - \frac{L_i}{2} \right)\left\|\Q_i^{(r,t+1)}-\Q_i^{(r,t)}\right\|_F^2.
\end{aligned}
\end{equation}

For the $\G$-subproblem, we have
\begin{align*}
u_G({\bm G}^{(r+1)}; \G^{(r)},\Q^{(r+1)})&\leq u_G({\bm G}^{(r)}; \G^{(r)},\Q^{(r+1)}) \\
                                           &= F({\bm G}^{(r)},\Q^{(r+1)})
\end{align*}
and thus
\[  F({\bm G}^{(r+1)},\Q^{(r+1)})+ \omega \|{\bm G}^{(r+1)}-{\bm G}^{r}\|_F^2 \leq  F({\bm Q}^{(r+1)},\G^{(r)}),    \]
or, equivalently
\begin{equation}\label{eq:suff_g}
\begin{aligned}
& F(\Q^{(r+1)},\G^{(r+1)}) - F(\Q^{(r+1)},\G^{(r)}) \\
& \leq -\omega\left\|\G^{(r+1)}-\G^{(r)}\right\|_F^2,\quad \forall \G^T\G ={\bm I},
\end{aligned}
\end{equation}
where $\omega=\left( \frac{I}{2{\gamma}} - \frac{I}{2} \right)>0$ if $\gamma<1$.
Combining \eqref{eq:suff_q} and \eqref{eq:suff_g}, we have
\begin{equation}\label{eq:suff}
\begin{aligned}
&F(\Q^{(r)},\G^{(r)}) - F(\Q^{(r+1)},\G^{(r+1)}) \\ &\geq\left( \frac{I}{2{\gamma}} - \frac{I}{2} \right)\left\|\G^{(r+1)}-\G^{(r)}\right\|_F^2\\
& +\sum_{t=0}^{T-1}\sum_{i=1}^I\left( \frac{1}{2\alpha_i} - \frac{L_i}{2} \right)\left\|\Q_i^{(r,t+1)}-\Q_i^{(r,t)}\right\|_F^2.
\end{aligned}
\end{equation}

Summing up $F(\Q^{(r)},\G^{(r)})$ over $r=0,1,\ldots,J-1$, we have
the following:
\begin{align}\label{eq:Zr}
&F(\Q^{(r)},\G^{(r)}) -F(\Q^{(r+1)},\G^{(r+1)}) \nonumber\\
&\geq \sum_{r=0}^{J-1} \omega\left\|\G^{(r+1)}-\G^{(r)}\right\|_F^2\nonumber\\
& + \sum_{r=0}^{J-1}\sum_{t=0}^{T-1}\sum_{i=1}^I\left( \frac{1}{2\alpha_i} - \frac{L_i}{2} \right)\left\|\Q_i^{(r,t+1)}-\Q_i^{(r,t)}\right\|_F^2.\nonumber\\
& = \sum_{r=0}^{J-1} \omega{\gamma}^2\left\|\G^{(r)} -\frac{\sum_{i=1}^I\X_i\Q_i^{(r+1)}}{I}+\G^{(r+1)}{\bm \Lambda}^{(r+1)}\right\|_F^2 \nonumber\\
&+  \sum_{r=0}^{J-1}\sum_{i=1}^I\sum_{t=0}^{T-1}\left( \frac{1}{2\alpha_i} - \frac{L_i}{2} \right) \alpha_{i}^2\left\|\tilde\nabla_{\Q_i}F(\Q^{(r,t)},\G^{(r)}) \right\|_F^2\nonumber\\
&\geq  \sum_{r=0}^{J-1} c Z^{(r+1)},
\end{align}
where
$c = \min\{ \omega\tilde{\gamma}^2, \{( \frac{1}{2\alpha_i} - \frac{L_i}{2} ) \alpha_i^2\}_{i=1,\ldots,I} \}.$
By the definition of $J$, we have
\begin{align*}
&\frac{F(\Q^{(0)},\G^{(0)}) - F(\Q^{(J)},\G^{(J)})}{J-1}\geq \frac{\sum_{r=0}^{J-1} c Z^{(r+1)}}{J-1}\geq c \cdot \delta\\
&\quad\Rightarrow  \delta \leq \frac{1}{c} \frac{F(\Q^{(0)},\G^{(0)}) - \bar{F}}{J-1}\Rightarrow  \delta \leq \frac{v}{J-1},
\end{align*}
\color{black}
where $\bar{F}$ is the lower bound of the cost function and
$v = \nicefrac{(F(\Q^{(0)},\G^{(0)}) - \bar{F})}{c}.$
This completes the proof.

\section{Proof of Theorem~\ref{thm:main}}
First consider an easier case where $\epsilon^{(r)}=0$ for all $r$.
Then, we have
${\bm Q}_i^{(r+1)}=({\bm X}_i^T{\bm X}_i+\mu_i{\bm I})^{-1}{\bm X}_i^T{\bm G}^{(r)}.$
Therefore, the update w.r.t. ${\bm G}$ is simply
to apply SVD on $\sum_{i=1}^I{\bm X}_i{\Q_i}/I = {\bm M}{\bm G}^{(r)}/I$.
In other words, there exists an invertible ${\bm \Theta}^{(r+1)}$ such that
\begin{equation}\label{eq:orthogonal}
  {\bm G}^{(r+1)}{\bm \Theta}^{(r+1)} = {\bm M}{\bm G}^{(r)},
\end{equation}
where ${\bm M}=\sum_{i=1}^I \X_i\X_i^\dag$ as before,
since the SVD procedure is nothing but a change of bases.
The update rule in \eqref{eq:orthogonal}, is essentially the \emph{orthogonal iteration} algorithm in \cite{GHGolub1996}.
Invoking \cite[Theorem 8.2.2]{GHGolub1996}, one can show that $\|{\bm U}_2^T{\bm G}^{(r)}\|_2$ approaches zero linearly.

The proof of the case where $\epsilon^{(r)}>0$ can be considered as an extension of round-off error analysis
of orthogonal iterations, and can be shown following the insights of \cite{lu2014large} and \cite{ge2016efficient} with proper modifications to accommodate the MAX-VAR GCCA case.
At the $r$th iteration, ideally, we have
${\tilde{\bm Q}_i^{(r+1)}}= ({\bm X}_i^T{\bm X}_i+\mu_i{\bm I})^{-1}{\bm X}_i^T{\bm G}^{(r)}$
if the $\Q$-subproblem is solved to optimality.
In practice, what we have is an inexact solution, i.e.,
\[{\Q}_i^{(r+1)}= ({\bm X}_i^T{\bm X}_i+\mu_i{\bm I})^{-1}{\bm X}_i^T{\bm G}^{(r)} + {\bm W}_i^{(r)},\]
where we have assumed that the largest singular value of ${\bm W}_i^{(r)}$ is bounded by $\epsilon$, i.e., $\|{\bm W}_i^{(r)}\|_2\leq\epsilon$.
Hence, one can see that
\[\sum_{i=1}^I{\bm X}_i\Q_i^{(r+1)} = \M\G^{(r)} + \sum_{i=1}^I \X_i{\bm W}_i^{(r)}.\]
Therefore, following the same reason of obtaining \eqref{eq:orthogonal}, we have
\[\G^{(r+1)}{\bm \Theta}^{(r+1)} =   \left(\M\G^{(r)} + \sum_{i=1}^I \X_i{\bm W}_i^{(r)}\right) ,\]
where ${\bm \Theta}^{(r+1)}\in\mathbb{R}^{K\times K}$ is a full-rank matrix since the solution via SVD is a change of  bases.
Consequently, we have
\begin{equation*}
\begin{bmatrix}
\U_1^T\G^{(r+1)}\\ \U_2^T\G^{(r+1)}
\end{bmatrix}{\bm \Theta}^{(r+1)}
=
\begin{bmatrix}
\bm \Lambda_1 \U_1^T\G^{(r)} + \U_1^T  \sum_{i=1}^I \X_i{\bm W}_i^{(r)} \\ \bm \Lambda_2 \U_2^T\G^{(r)} + \U_2^T  \sum_{i=1}^I \X_i{\bm W}_i^{(r)}
\end{bmatrix}
.
\end{equation*}
Now, we denote
 \[ \Delta_1^{(r)} =   \U_1^T  \sum_{i=1}^I \X_i{\bm W}_i^{(r)} ,\quad   \Delta_2^{(r)} =   \U_2^T  \sum_{i=1}^I \X_i{\bm W}_i^{(r)},    \]
as two error terms at the $r$th iteration. 
Next, let us consider the following chain of inequalities:
\begin{align}
 &\left\| \U_2^T\G^{(r+1)}\left( \U_1^T\G^{(r+1)} \right)^{-1}  \right\|_2\\
 & = \left\|\left(\bm \Lambda_2 \U_2^T\G^{(r)} + \Delta_2^{(r)}  \right)\left(\bm \Lambda_1 \U_1^T\G^{(r)} + \Delta_1^{(r)}   \right)^{-1}\right\|_2 \nonumber\\
 &\leq \frac{ \left\|\left(\bm \Lambda_2 \U_2^T\G^{(r)} + \Delta_2^{(r)}  \right)(\U_1^T\G^{(r)})^{-1}\right\|_2}{\sigma_K\left(\bm \Lambda_1  + \Delta_1^{(r)}(\U_1^T\G^{(r)})^{-1}   \right)} \nonumber\\
 &\leq \frac{\lambda_{K+1}\left\| \U_2^T\G^{(r)}\left( \U_1^T\G^{(r)} \right)^{-1}  \right\|_2 + {\|\Delta_2^{(r)}\left( \U_1^T\G^{(r)} \right)^{-1}\|_2}}{\lambda_K -  \|\Delta_1^{(r)}\left( \U_1^T\G^{(r)} \right)^{-1}\|_2} \nonumber \\
 & \leq \left\| \U_2^T\G^{(r)}\left( \U_1^T\G^{(r)} \right)^{-1}  \right\|_2 \left(\frac{\lambda_{K+1} + \frac{\|\Delta_2^{(r)}\|_2}{\sigma_{\min}( \U_2^T\G^{(r)} )}}{\lambda_K -  \frac{\|\Delta_1^{(r)}\|_2}{\sigma_{\max}\left( \U_1^T\G^{(r)} \right)}}\right) \label{eq:key_upper_bound}.
\end{align}
Assume that the following holds:
\begin{align} \label{eq:suff_small}
 &\max\{\|\Delta_1^{(r)}\|_2,\|\Delta_2^{(r)}\|_2\}\\
 &\leq \frac{\lambda_K - \lambda_{K+1}}{3}\min\left\{\sigma_{\min}\left( \U_2^T\G^{(r)} \right),\sigma_{\max}\left( \U_1^T\G^{(r)} \right)\right\}.\nonumber
\end{align}
Then, one can easily show that
\begin{align}\label{eq:key_ineq}
 &\left\| \U_2^T\G^{(r+1)}\left( \U_1^T\G^{(r+1)} \right)^{-1}  \right\|_2 \nonumber \\
 &\quad\quad \leq \varrho \left\| \U_2^T\G^{(r)}\left( \U_1^T\G^{(r)} \right)^{-1}  \right\|_2 
\end{align}
where 
\[ \varrho = \left( \frac{2\lambda_{K+1}+\lambda_K}{2\lambda_K+\lambda_{K+1}} \right)<1. \]

One can see that
\begin{align}\label{eq:key_ineq_2}
\left\| \U_2^T\G^{(r+1)}\right\|_2&\leq\left\| \U_2^T\G^{(r+1)}\left( \U_1^T\G^{(r+1)} \right)^{-1}  \right\|_2 \nonumber\\
&\leq \varrho^r \left\|( \U_2^T\G^{(0)} )\left( \U_1^T\G^{(0)}\right)^{-1} \right\|_2 \nonumber\\
&\leq \varrho^r \tan(\theta),
\end{align}
where the first inequality holds because of $\|\U_1^T\G^{(r+1)}\|_2\leq 1$.
By noticing that $\|\U_2^T\G^{(0)} \|_2=\sin(\theta)$ and $\|( \U_1^T\G^{(0)})^{-1}  \|_2= 1/\cos(\theta)$ \cite[Theorem 8.2.2]{GHGolub1996}, we obtain the last inequality.

In addition, we notice that 
\[\max\{\|\Delta_1^{(r)}\|_2,\|\Delta_2^{(r)}\|_2\} \leq \sum_{i=1}^I\lambda_{\max}(\X_i)\epsilon^{(r)}.\]
This means that to ensure linear convergence to a global minimal solution, it suffices to have
\begin{align}\label{eq:eps}
\epsilon^{(r)} &\leq \frac{\lambda_K - \lambda_{K+1}}{3\sum_{i=1}^I\lambda_{\max}(\X_i)} \\
&\times \min\left\{\sigma_{\min}\left( \U_2^T\G^{(r)} \right),\sigma_{\max}\left( \U_1^T\G^{(r)} \right)\right\}  \nonumber 
\end{align}
in the worst case. 

The last piece of the proof is to show that $(\U_1^T\G^{(r)})^{-1}$ in \eqref{eq:key_upper_bound} always exists. 
Note that if  \eqref{eq:key_ineq_2} holds, then $\U_1^T\G^{(r)}$ is always invertible under the condition stated in \eqref{eq:sigma}. The reason is that we always have \cite{GHGolub1996}
\[\sigma_{\max}^2(\U_2^T\G^{(r)})+\sigma_{\min}^2(\U_1^T\G^{(r)})=1.\]
Therefore, $\sigma_{\min}^2(\U_1^T\G^{(r)})$ monotonically increases since $\sigma_{\max}(\U_2^T\G^{(r)})$ decreases when \eqref{eq:eps} (and thus \eqref{eq:key_ineq_2}) holds.
Hence, if $\sigma_{\min}(\U_1^T\G^{(0)})>0$, we have $\sigma_{\min}(\U_1^T\G^{(r)})>0$ for all $r>1$.

\color{black}



\end{document}